\documentclass[10pt,twocolumn,letterpaper]{article}

\usepackage[pagenumbers]{iccv} 

\usepackage[dvipsnames]{xcolor}

\definecolor{iccvblue}{rgb}{0.21,0.49,0.74}
\usepackage[pagebackref,breaklinks,colorlinks,allcolors=iccvblue]{hyperref}

\def\paperID{779} 
\def\confName{ICCV}
\def\confYear{2025}
\usepackage{multirow}
\usepackage{float}
\usepackage{makecell}
\usepackage{pifont}
\usepackage{xcolor}
\definecolor{cvprblue}{rgb}{0.21,0.49,0.74}
\definecolor{building}{RGB}{200, 200, 20}
\definecolor{pond}{RGB}{0, 170, 240}
\definecolor{forest}{RGB}{0, 200, 80}
\definecolor{land}{RGB}{200, 90, 20}

\usepackage[accsupp]{axessibility}  

\title{SkySense V2: A Unified Foundation Model for Multi-modal Remote Sensing}

\author{Yingying Zhang$^1$\quad Lixiang Ru$^1$\quad Kang Wu$^2$\quad Lei Yu$^1$\quad Lei Liang$^1$\quad Yansheng Li$^2$\quad Jingdong Chen$^1$\\$^1$Ant Group \quad $^2$ Wuhan University\\{\tt qichu.zyy@antgroup.com}}

\begin{document}
\maketitle
\begin{abstract}
The multi-modal remote sensing foundation model (MM-RSFM) has significantly advanced various Earth observation tasks, such as urban planning, environmental monitoring, and natural disaster management. However, most existing approaches generally require the training of separate backbone networks for each data modality, leading to redundancy and inefficient parameter utilization. Moreover, prevalent pre-training methods typically apply self-supervised learning (SSL) techniques from natural images without adequately accommodating the characteristics of remote sensing (RS) images, such as the complicated semantic distribution within a single RS image. In this work, we present SkySense V2, a unified MM-RSFM that employs a single transformer backbone to handle multiple modalities. This backbone is pre-trained with a novel SSL strategy tailored to the distinct traits of RS data. In particular, SkySense V2 incorporates an innovative adaptive patch merging module and learnable modality prompt tokens to address challenges related to varying resolutions and limited feature diversity across modalities. In additional, we incorporate the mixture of experts (MoE) module to further enhance the performance of the foundation model. SkySense V2 demonstrates impressive generalization abilities through an extensive evaluation involving 16 datasets over 7 tasks, outperforming SkySense by an average of 1.8 points.

\end{abstract}
\begin{figure}
	\centering
		\includegraphics[scale=.3]{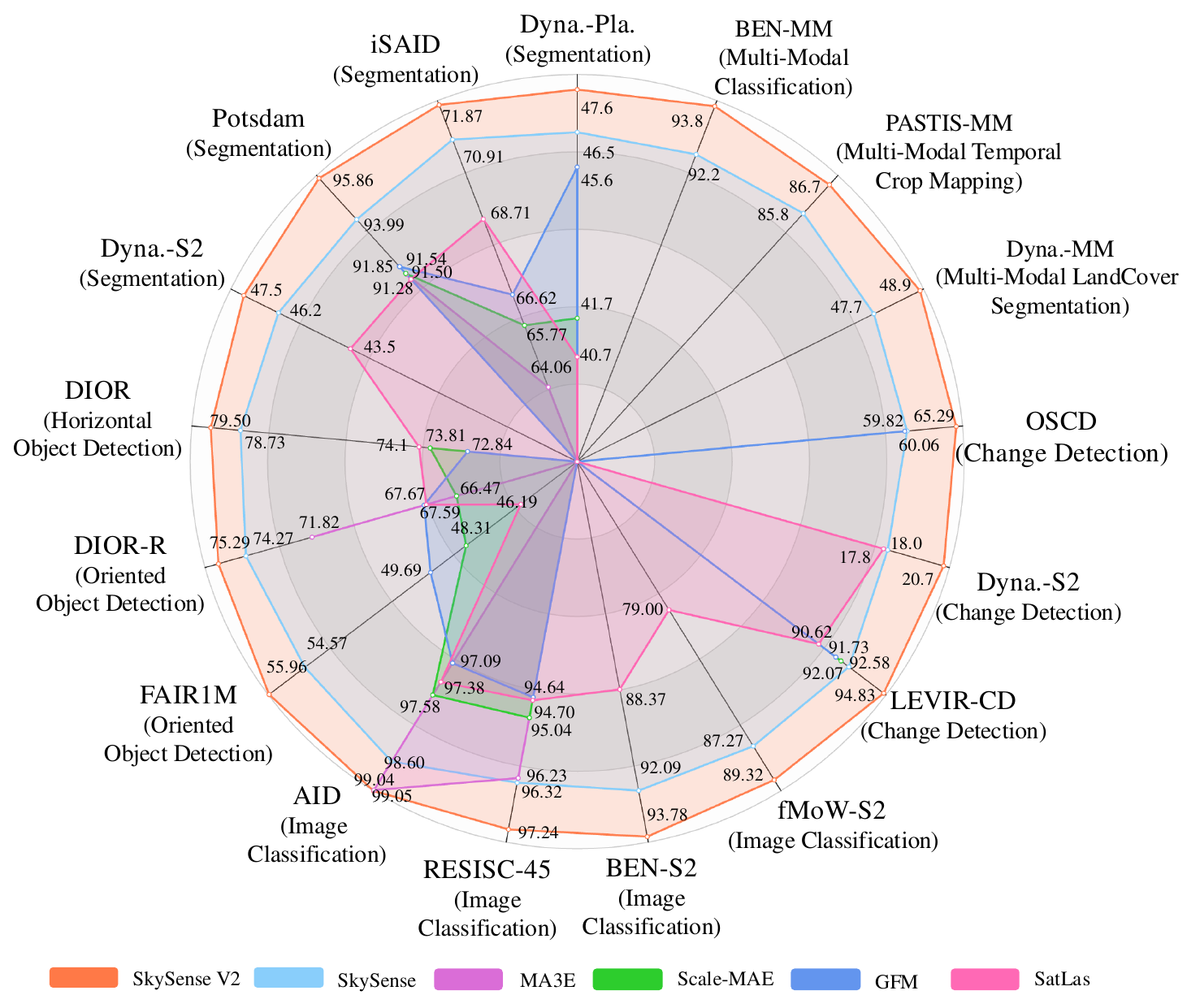}
	    \caption{SkySense V2 has achieved superior performance on 16 datasets over 7 distinct tasks compared with its predecessor SkySense and supports a board range of tasks.}
	\label{all_res}
\vspace{-15pt}
\end{figure}

\vspace{-10pt}
\section{Introduction}
\label{sec:intro}
Advancements in multi-modal remote sensing foundation models (MM-RSFM) have significantly enhanced the utilization of remote sensing (RS) data across numerous Earth observation (EO) applications\cite{yuan2021review,cheng2016survey,lv2022land,cao2023multi}, such as environmental monitoring, agriculture management, natural disaster response, land cover (or use) mapping \etc. Recently, Guo \etal introduced SkySense \cite{SkySense}, the largest MM-RSFM to date. SkySense has demonstrated impressive generalization capabilities through extensive evaluations across many different tasks. A key factor in its success is the simultaneous training of geographically aligned multi-modal and multi-temporal RS data. To accommodate the varying resolutions of different modalities, SkySense employs the Swin Transformer \cite{liu2021swin} for processing high-resolution (HR) optical images, while using the Vision Transformer (ViT) \cite{dosovitskiy2020image} for medium-resolution multi-spectral (MS) data and synthetic aperture radar (SAR) data. However, this separate backbone design results in redundancy and inefficient use of parameters. On the other hand, SkySense's pre-training approach is primarily based on DINOv2 \cite{oquab2023dinov2}, renowned for its efficiency in self-supervised learning (SSL). Although it has demonstrated strong performance with RS images, there remains considerable room for improvement by adequately accommodating the involved semantic distribution within a single RS image. 

In this paper, we introduce SkySense V2, an advanced MM-RSFM pre-trained through a novel SSL approach tailored to the distinctive traits of RS data. To be specific, SkySense V2 leverages a unified transformer backbone capable of processing data across diverse modalities and resolutions. It is pre-trained using a new SSL strategy that utilizes query-based attention to gather the similar semantic features distributed in different regions of RS images.

Designing a unified backbone for geographically aligned multi-modal data presents two primary challenges. The first is processing feature resolutions across data with varying ground sample distance (GSD) during simultaneous pre-training of multiple modalities. To address this challenge, we introduce an innovative Adaptive Patch Merging (APM) module, which is integrated after each stage of the unified backbone. This module determines whether to reduce feature resolution according to the specific feature resolution requirements of each modality. For instance, when dealing with HR optical images, the APM module activates resolution reduction at each stage. In contrast, for medium-resolution data, such as MS and SAR data, the feature resolution is preserved throughout the stages. This design enables the backbone to efficiently process data of different resolutions while adapting output feature resolutions according to actual requirements, which is crucial for the joint training and fusion of multi-modal features. 

The second challenge is that fully sharing parameters across the different modalities can reduce the feature diversity. To address this, inspired by the work of \cite{ViTR} and \cite{VPT}, we introduce learnable modality prompt tokens for each modality. By interacting with these individual modal prompt tokens through an attention mechanism, the pre-trained model can capture the unique characteristics of each modality better. Compared to SkySense, our unified design significantly enhances the efficiency of parameter utilization. To be specific, SkySense uses three separate backbones, Swin-H for HR optical data, ViT-L for MS data, and ViT-L for SAR data, collectively totaling 1.26 billion parameters. In contrast, SkySense V2's unified transformer backbone is capable of simultaneously processing all three modalities while utilizing only 665 million parameters. This efficient use of parameters not only optimizes the architecture but also allows for potential model scaling to further boost performance. Therefore, we integrate the mixture of experts module \cite{MOE}, a technique commonly used in various large language models \cite{SwitchTS, TMoEDS, MOE_IT} to further enhance the performance of the foundation model.

\begin{figure}[!htp]
    \centering  
   \begin{subfigure}{0.49\linewidth} 
      \centering   
      \includegraphics[width=1\linewidth]{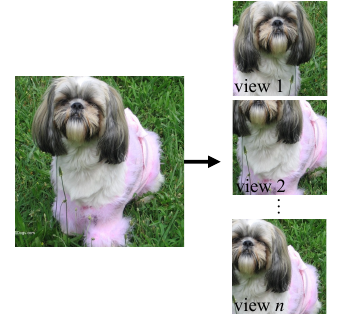}
        \caption{Views of natural image}
        \label{fig:view_cmp_n}
    \end{subfigure} 
    \begin{subfigure}{0.49\linewidth} 
      \centering   
      \includegraphics[width=1\linewidth]{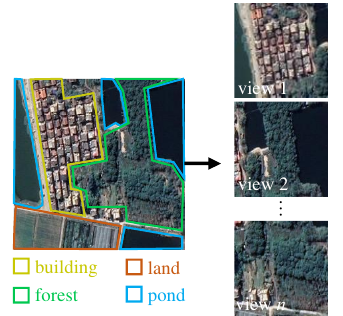}
        \caption{Views of RS image}
        \label{fig:view_cmp_rs}
    \end{subfigure} 
    \caption{Comparison of different augmented views from natural and  RS images in SSL. Natural image views often concentrate on a singular subject, such as a "dog". RS image views encompass a variety of subjects, including building, forest, pond, and land.}
    \label{fig:view_cmp}
    \vspace{-10pt}
\end{figure}

To effectively adapt SSL techniques developed for natural images to the pre-training of RSFM, it is essential to consider the distinct differences in data characteristics. Natural images typically feature a single, clear subject, such as a cat, dog, or person. In contrast, RS images comprise multiple subjects distributed across various regions of the image. For instance, as illustrated in Figure \ref{fig:view_cmp}, one section of an RS image may show a \textcolor{building}{building}, while other areas might show a \textcolor{forest}{forest}, \textcolor{pond}{pond}, or  \textcolor{land}{land}. Current SSL methods perform contrastive learning directly between different views using various augmented crops, which is effective for natural images due to their singular focus. However, when applied to RS images, this approach can result in semantic inaccuracies, as different views may capture different subjects. For example, in Figure \ref{fig:view_cmp_rs}, view 1 of the RS image primarily contains \textcolor{building}{building}, whereas view 2 shows \textcolor{forest}{forest} and \textcolor{pond}{pond}.

To tackle this challenge, we propose a novel approach called Query-based Semantic Aggregation Contrastive Learning (QSACL). This method employs multiple learnable queries to perform cross-attention with features from different views, generating semantically aggregated features. We then apply contrastive learning to these aggregated features pairs derived from the same query. As demonstrated in Figure \ref{qsacl_attn} from our ablation study, utilizing different queries can aggregate consistent semantic features across multiple views, which enhances the accuracy of contrastive learning.

We evaluated SkySense V2 on a diverse set of 16 datasets \cite{waqas2019isaid,toker2022dynamicearthnet,li2020object,cheng2022anchor,sun2022fair1m,xia2017aid,cheng2017remote,sumbul2020bigearthnet,cong2022satmae,chen2020spatial,daudt2018urban,garnot2022multi}, covering a range of task types, modalities, and spatial scales. As illustrated in Figure~\ref{all_res}, SkySense V2 demonstrates substantial improvements in performance over its predecessor, SkySense, achieving state-of-the-art (SOTA) results across various modalities of EO tasks. The experimental results across all test scenarios highlight its competitive advantage over existing RSFMs in a wide array of EO interpretation tasks.

\section{Related Work}
\subsection{Remote Sensing Foundation Model}
RSFMs are characterized by their ability to leverage vast amounts of data through self-supervised learning (SSL) technologies, enabling them to learn robust feature representations without the need for extensive annotations. The predominant RSFMs utilize Contrastive Learning (CL) or Masked Image Modeling (MIM). Typical works in CL include RS-BYOL \cite{RS_BYOL}, GASSL \cite{GASSL}, DINO-MC \cite{wanyan2023dino}, SeCo \cite{manas2021seasonal}, and CACo \cite{mall2023change}. Other research efforts focus on enhancing MIM framework, such as RingMo \cite{sun2022ringmo}, S2MAE \cite{Li_2024_CVPR}, MA3E \cite{MA3E}, SatMAE \cite{cong2022satmae}, and SatMAE++ \cite{SatMAE++}. Additionally, studies like CMID \cite{muhtar2023cmid} and GFM \cite{mendieta2023gfm} explore the intersection of CL and MIM through a self-distillation approach.
Recently, Guo \etal introduced a comprehensive MM-RSFM known as SkySense \cite{SkySense}, which features a factorized multi-modal spatiotemporal encoder. This architecture facilitates independent spatial feature extraction and multi-modal temporal fusion. As the largest MM-RSFM to date, SkySense exhibits exceptional generalization capabilities across a wide range of RS datasets. However, it does face a challenge: the inefficient utilization of parameters due to the separate backbone designs for different modalities. In this work, we propose SkySense V2 to address this issue by employing a unified transformer backbone design.
\subsection{Unified Framework for Multi-modal Learning}
Multi-modal learning aims to train models that effectively process and relate information from various modalities. Substantial advancements have been made in this field, culminating in numerous studies focused on creating unified networks for the integration and processing of diverse modalities. The main work includes VLMO \cite{VLMO}, Meta-Transformer \cite{Meta_trans}, Uni-Perceiver \cite{Uni_Perceiver}, and UniTR \cite{UniTR}, \etc. There are several methods that utilize a unified backbone design for multi-modal RS data. For example, OFA-Net \cite{ofa_net} pre-trains a single transformer backbone on a curated multi-modal dataset using MIM. Han \etal proposed a RSFM called msGFM\cite{msGFM}, which utilizes a shared transformer encoder while employing different decoders for each modality. In contrast to our approach, both OFA-Net and msGFM simply share all of parameters of backbone and learn feature representations for multi-modal data separately, which results in a lack of geographical alignment and fusion between features from different modalities. AnySat \cite{AnySat} employs a joint embedding predictive architecture for a multi-modal model, training a single model on heterogeneous data in a SSL manner. However, the backbone design and SSL pre-training methods of our SkySense V2 differ significantly from AnySat.


\section{SkySense V2} 
\subsection{Model Architecture}
\subsubsection{Unified Transformer Backbone}
\begin{figure}
	\centering
		\includegraphics[scale=.34]{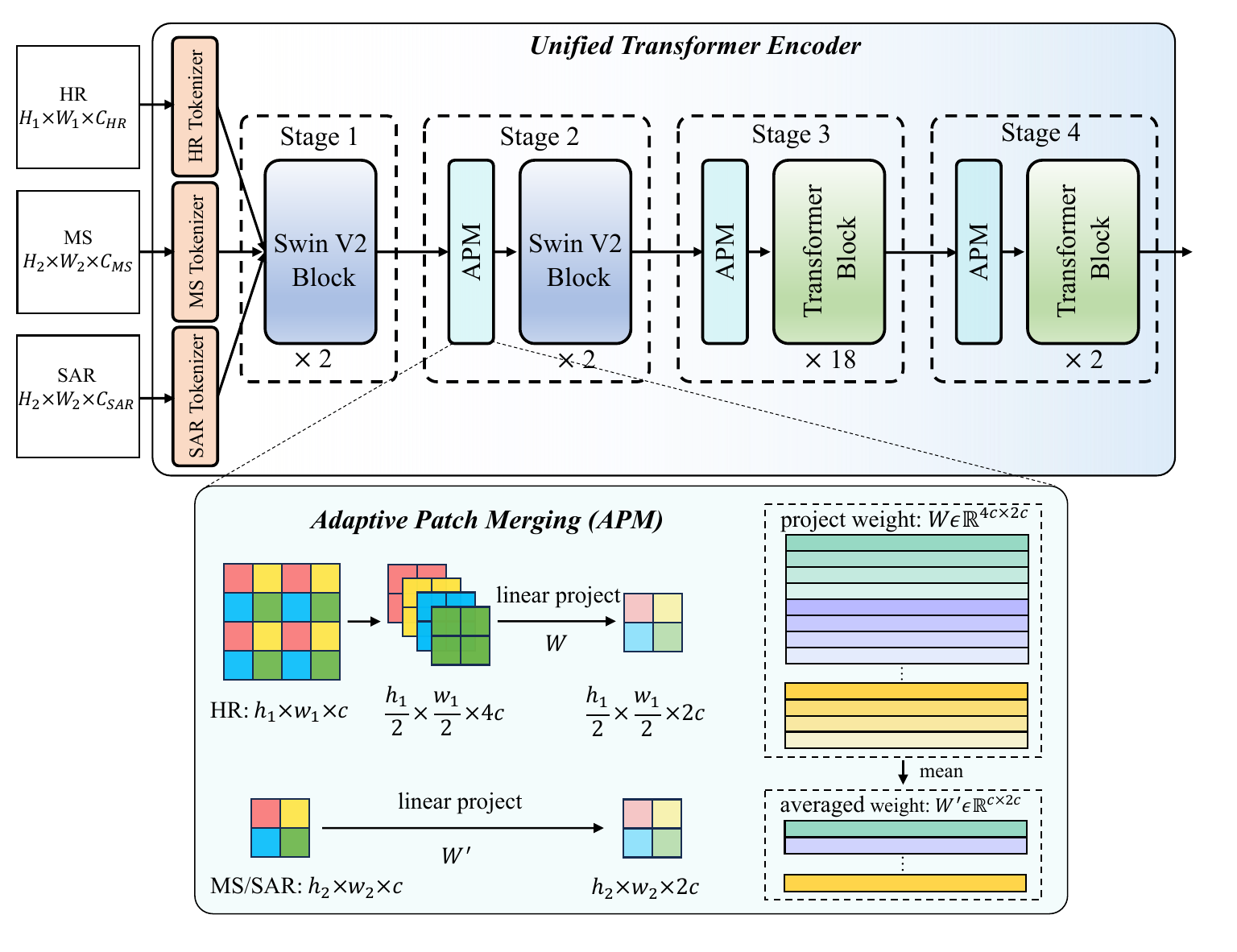}
	    \caption{Overview of the unified transformer backbone model in SkySense V2: The entire backbone shares all parameters across different modalities, with the exception of distinct tokenizers.}
	\label{unified_bacbone}
    \vspace{-10pt}
\end{figure}

As illustrated in Figure \ref{unified_bacbone}, our unified transformer backbone is a hierarchical encoder structure with four stages. In the first two stages, we employ Swin Transformer V2 Blocks (SwinV2B) \cite{Swinv2} to incorporate essential visual priors, such as locality and translation invariance. The window-based self-attention mechanisms in SwinV2B also help reduce computational complexity compared to global self-attention, particularly given the high spatial resolution of features in these initial stages. In the last two stages, we utilize vanilla Transformer Blocks (TB) \cite{dosovitskiy2020image} for two main reasons: First, the spatial resolution of features in these stages is relatively low, making the computational costs of global self-attention more manageable. Second, our ablation study demonstrates that global self-attention can be complementary to window-based self-attention, enabling the model to achieve stronger representations. 

Given a group of multi-modal inputs consisting of a high-resolution optical image $x_{HR}$, multi-spectral data $x_{MS}$, and synthetic aperture radar data $x_{SAR}$ where each 'pixel' in the different modalities is naturally aligned by geo-location, we employ three distinct tokenizers to process these inputs into tokens. In each tokenizer, the input data is first divided into non-overlapping $4 \times 4$ patches. Subsequently, a linear embedding layer is applied to this raw, patched data to project it into patch tokens. Subsequently, four stages of either SwinV2B or TB, with shared parameters across modalities, are applied to these patch tokens.

Due to the varying GSD of satellite sensors across different modalities, input data corresponding to the same region at specific geo-locations during joint aligned training exhibit differing spatial resolutions. To harmonize these spatial resolutions for tokens from various modalities, we propose Adaptive Patch Merging (APM) and utilize it to selectively reduce spatial resolution at each stage, except for Stage 1. Specifically, for the tokens of optical image data with high spatial resolution, APM concatenates the features from groups of $2 \times 2$ neighboring patches and applies a linear layer to the $4c$-dimensional concatenated features. This process reduces the number of tokens by a factor of $2\times2=4$ downsampling of spatial resolution, while the output dimension is set to $2c$. In contrast, for the tokens of MS and SAR data with lower spatial resolution, APM maintains the resolution by applying a linear projection with averaging the weights across the input dimensions. By integrating the APM module, our unified backbone can efficiently process multi-modal data with varying resolutions while maintaining the spatial alignment of features across modalities

\subsubsection{Modality-specific Prompt Tokens}

\begin{figure}
	\centering
		\includegraphics[width=0.5\textwidth]{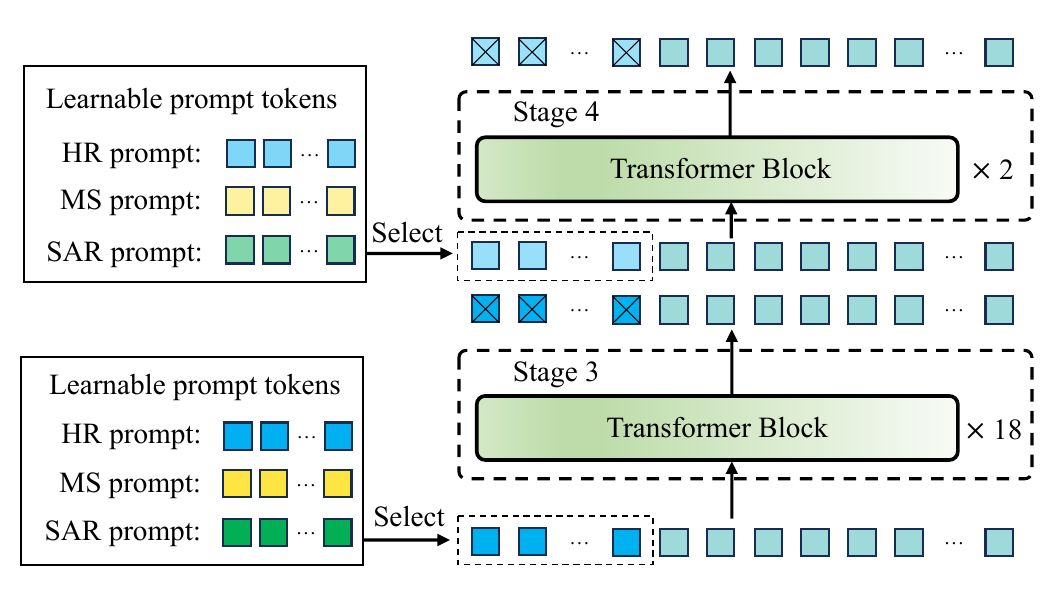}
	    \caption{The illustration of modality-specific prompt tokens added to the unified backbone.}
	\label{MsPT}
    \vspace{-10pt}
\end{figure}

Our unified transformer backbone utilizes fully shared parameters across different modalities. To enhance feature diversity, we introduce learnable modality-specific prompt tokens for each modality. By interacting with these individual modal prompt tokens through an attention mechanism, the pre-trained model can better capture the unique characteristics of each modality. As illustrated in Figure \ref{MsPT}, we incorporate $N$ learnable prompt tokens for each modality in the last two stages. We denote the input tokens of each stage as $E_{i}^{j} \in \mathbb{R}^{h_{j}w_{j}\times c_{j}}, i \in \{HR, MS, SAR \}, j \in \{3, 4\}$, where $h_j$ and $w_j$ are the height and width of the spatial resolution at stage $j$, while $c_j$ refers to the dimension of the tokens. For each modality $i$, we insert the modality-specific learnable prompt tokens $P^{j}_{i} \in \mathbb{R}^{N \times c_{j}}$ into the input of the first block in $j$-th stage $\mathcal{F}_j$. At the last block of each stage, these prompt tokens are discarded. The entire process can be formulated as:
\begin{equation}
\begin{aligned}
&[E_{drop}, E_i^4] = \mathcal{F}_3 ([P^3_{i}, E_i^3]),  \\
&[E_{drop}, E_i^{out}] = \mathcal{F}_4 ([P^4_{i}, E_i^4]).
\end{aligned}
\end{equation}
Here, $E_{drop}$ denotes the discarded tokens, and $E_i^{out}$ represents the final output tokens of backbone. Our design of modality-specific prompts enhances feature diversity while maintaining full parameter sharing by incorporating only a small number of modality-specific parameters. 

\begin{figure*}
	\centering
		\includegraphics[width=1.0\textwidth]{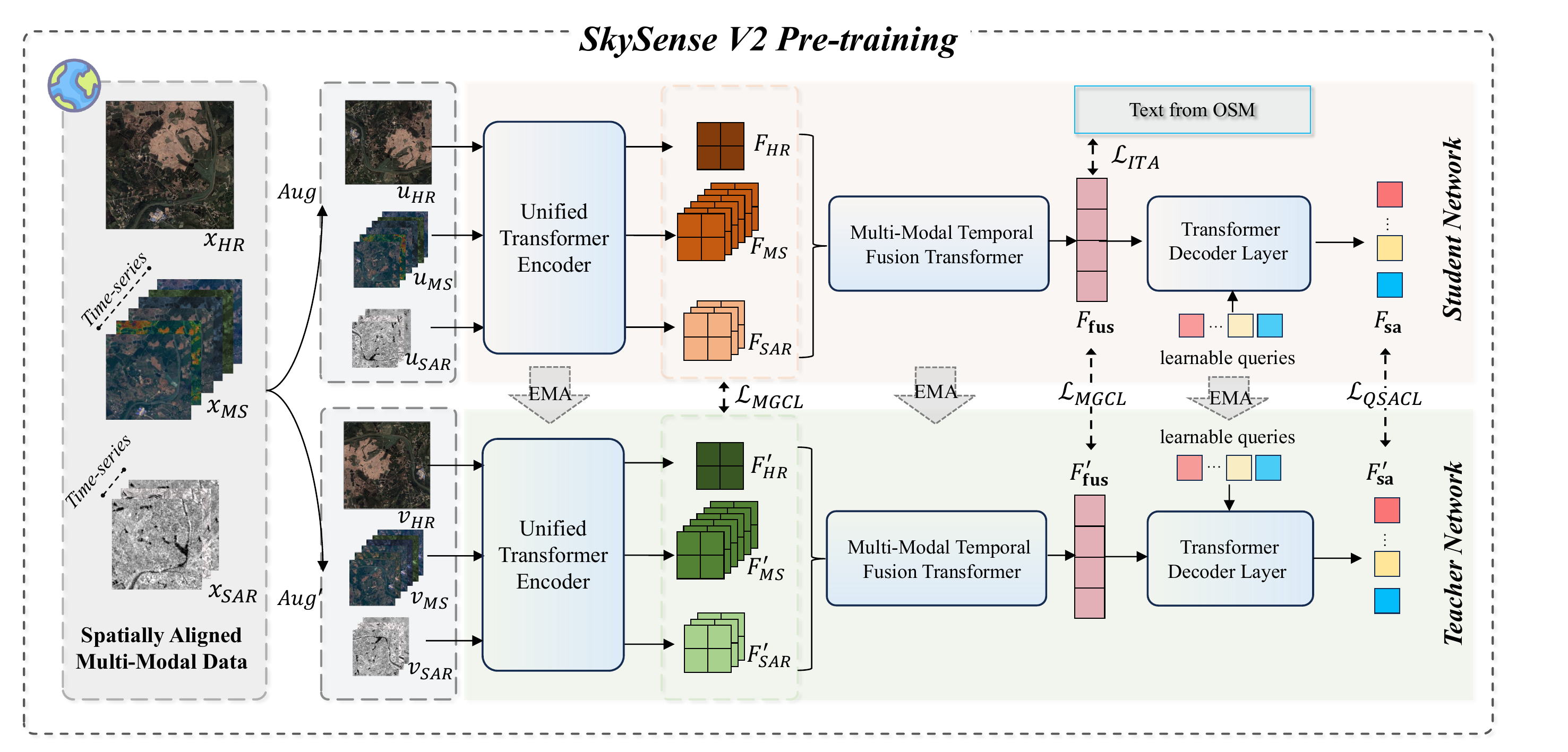}
	    \caption{Overview of the SkySense V2 pre-training pipeline. The SkySense V2 employs data augmentations on multi-modal inputs, which are then fed into both the student and teacher networks. The pre-training process incorporates Multi-Granularity Contrastive Learning, Dense Image-Text Alignment, and Query-based Semantic Aggregation Contrastive Learning to effectively train the network. Once pre-training is complete, the parameters from the teacher branch are utilized for downstream applications.} 
	\label{pipeline}
\end{figure*}

\subsubsection{Scaling Up Model Capacity}
Our unified backbone design across various modalities significantly enhances parameter utilization efficiency. To further boost the model’s capabilities, we integrate a Mixture of Experts (MoE) \cite{MOE} into the backbone network. We choose MoE instead of simply increasing the backbone's width and depth because MoE enables models to be pre-trained with substantially less computation by leveraging sparse feed-forward layers (\ie, experts) for individual tokens. Following common approaches \cite{ResidualMoE,Task-customizedMOE}, we insert MoE modules into the last $L$ transformer blocks, replacing the original feed-forward network (FFN) layers. Each MoE module consists of $M$ experts (denoted as $\mathcal{E}_i(\cdot), i = 1, 2, . . . , M$), which share the same structure as the FFN and operate as independent networks. For the gating network in MoE, we employ a learnable linear layer followed by a Softmax function, $\mathcal{G}(x)=Softmax(Wx)$ , where W is the gating parameter. Finally, the output of the MoE is computed as a linear combination of the outputs from the selected experts, weighted by the corresponding gate values. This process is formulated as follows:
\begin{equation}
\begin{aligned}
&MOE(x) = \sum_{i \in \mathcal{T}} \mathcal{G}_i(x) \cdot \mathcal{E}_i(x),
\end{aligned}
\end{equation}
$\mathcal{T}$ represent the set of the top-$k$ indices. In our SkySense V2 backbone, we set $L=6, M=8$, and $k=1$.

\subsection{Overall Pre-training Pipeline}
Figure \ref{pipeline} provides an overview of our pre-training procedure. The pre-training framework of SkySense V2 primarily adopts the teacher-student architecture from SkySense \cite{SkySense}, where the teacher network's parameters are updated using an exponential moving average (EMA) \cite{oquab2023dinov2} of the student network's parameters. To train SkySense V2, we employ the Multi-Granularity Contrastive Learning (MGCL) loss $\mathcal{L}_{MGCL}$ 
as proposed in SkySense. For each modality, MGCL utilizes contrastive loss \cite{oquab2023dinov2} to align the representations of teacher and student networks at pixel, object and image-level granularity, respectively. Additionally, we incorporate the unsupervised Geo-Context Prototype Learning (GCPL), as validated in SkySense, to enhance the learning of complementary regional context clues, aiding downstream tasks. To further enhance dense interpretation capabilities, we introduce an auxiliary supervision loss $\mathcal{L}_{ITA}$ to employ dense Image-Text Alignment (ITA) according to OpenStreetMap (OSM) labels\footnote{https://www.openstreetmap.org/}. Detailed implementations of MGCL, GCPL, and ITA are included in the Appendix A. In following paragraph, we will focus on proposed query-based semantic aggregation contrastive learning (QSACL).

\begin{figure}[H]
    \vspace{-20pt}
	\centering
		\includegraphics[width=0.48\textwidth]{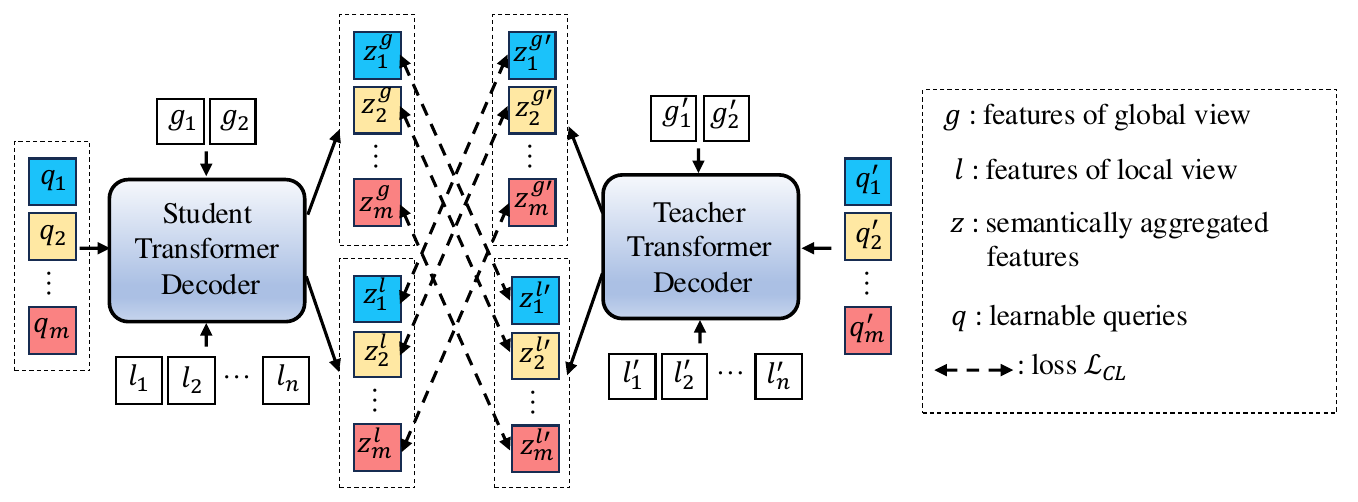}
	    \caption{Computation pipelines for QSACL with multi-crop augmentation (2 global and $n$ local views). }
	\label{qsacl}
\end{figure}

\paragraph{Query-based Semantic Aggregation Contrastive Learning.}
QSACL utilizes learnable queries to aggregate similar semantics across different regions of images, enabling more accurate contrastive learning. Given features $g_1, g_2$ from two global views and $l_1, l_2, ... l_n$ from $n$ local views (where $g$ and $l$ denotes the fused features  $F_{\mathbf{fus}}$ from global and local view), we employ a transformer decoder layer to perform cross-attention between $m$ learnable queries $q_1, q_2, ..., q_m$ and the feature sets $g$ and $l$. This process generates sets of globally and locally aggregated features set $z^g_i$ and  $z^l_i$ for each $i=1,2, ... m$. As is shown in Figure \ref{qsacl}, we use superscript $\prime$ denotes features from the teacher network, the QSACL loss $\mathcal{L}_{QSACL}$ is computed by averaging the loss $\mathcal{L}_{CL}$ of aggregation features between local and global views for each query:
\begin{equation}
\begin{aligned}
\mathcal{L}_{QSACL} = \frac{1}{2m} \sum_{i=1}^m (\mathcal{L}_{CL}(z_i^{g}, z_i^{l\prime}) + \mathcal{L}_{CL}(z_i^{l}, z_i^{g\prime})),
\end{aligned}
\label{equ:pixel}
\end{equation}
\begin{equation}
\begin{aligned}
\mathcal{L}_{CL} (x, x^{\prime}) = -\mathcal{H}(x) \log(\mathcal{H}^{\prime}(x^{\prime})).
\end{aligned}
\label{equ:pixel}
\end{equation}
Here, $\mathcal{H}$ and $\mathcal{H}^{\prime}$ denotes the learning head and corresponding EMA part as defined in\cite{caron2021emerging}. The overall training objective is the weighted sum of $\mathcal{L}_{MGCL}$, $\mathcal{L}_{ITA}$, and $\mathcal{L}_{QSACL}$, \ie, 
\begin{equation}
\begin{aligned}
\mathcal{L} = \lambda_1\mathcal{L}_{MGCL} +\lambda_2\mathcal{L}_{ITA}+\lambda_3\mathcal{L}_{QSACL}.
\end{aligned}
\end{equation}

\section{Experiments} 
Following SkySense \cite{SkySense}, we conducted experiments on 16 datasets across various modalities and tasks to thoroughly evaluate SkySense V2. The applications of pre-trained SkySense V2 to diverse tasks are detailed in Appendix B. Each module is engineered for combined or individual use, providing the flexibility to be frozen or fine-tuned as needed. 

\subsection{Pre-training Implementation}

\paragraph{Training Datasets.} We use the same training dataset of SkySense \cite{SkySense} to train SkySense V2. This extensive dataset comprises approximately 21 million multi-modal RS imagery sets. Each set comprises three components: a high-resolution optical RGB image of spatial dimension $2048\times 2048$, a sequence of Sentinel-2 (S2) multi-spectral (MS) images with 10 spectral bands and spatial dimensions of $64\times 64$ (average sequence length of 65 images), and a sequence of Sentinel-1 (S1) synthetic aperture radar (SAR) images with 2 bands, also having spatial dimensions of $64 \times 64$ (average sequence length of 10 images). In each training iteration, we randomly sample 10 S2 images and 5 S1 images of different time series to accommodate the GPU memory constraints.

\paragraph{Setting of Unified Backbone.} We initially set the token dimension to $C=352$. In the first two stages, the window size for the Swin Transformer V2 Blocks (SwinV2B) is configured to 8. Each head has a query dimension of 32, and each MLP layer consists of two full-connected layers with dimensions expanding by a factor of 4 consistently across all blocks within the unified backbone. Prompt tokens are incorporated during the last two stages, with the number of prompt tokens being the same for each modality. We allocate 4 prompt tokens for stage-3 and stage-4.

\paragraph{Pre-training Settings.} SkySense V2 is trained using a batch size of 1024, distributed across 128 H20 GPUs. The model undergoes a total of 600k iterations, utilizing the AdamW optimizer~\cite{loshchilov2018fixing}. The learning rate is initially set to $2 \times 10 ^ {-4}$ and decays to $1 \times 10 ^ {-6}$  following a cosine annealing schedule \cite{SGDRSG}. Additional implementation details can be found in the Appendix C.

\subsection{Single-Modal Tasks} We evaluate SkySense V2 on four common single-modal RS interpretation tasks: scene classification, object detection, semantic segmentation, and change detection. The experiments are conducted via fine-tuning a pre-trained SkySense V2 backbone along with a task-specific head. Further implementation details can be found in the Appendix D.

\paragraph{Scene Classification.}
In Table~\ref{tab:scene_classification}, we report the performance of scene classification on four widely used datasets: AID~\cite{xia2017aid} and RESISC-45~\cite{cheng2017remote}, both featuring static RGB optical images; BEN-S2~\cite{sumbul2020bigearthnet}, which includes static MS images; and fMoW-S2~\cite{cong2022satmae}, which contains temporal MS images. For each dataset, the performance is evaluated using different training ratios (TR) following the methodology \cite{cong2022satmae,manas2021seasonal}. In our experiments, we employ a simple linear classifier as the classification head. For multi-label datasets, we use the mean average precision (mAP) as the evaluation metric, whereas for single-label datasets, we report the overall accuracy (OA). As illustrated in Table~\ref{tab:scene_classification}, SkySense V2 consistently outperforms previous RSFMs across various datasets and training configurations, particularly in low TR settings. This demonstrates that SkySense V2 exhibits a stronger representation capability compared to other RSFMs, including its predecessor, SkySense \cite{SkySense}.

\begin{table}
\centering
\setlength\tabcolsep{4pt}%
\scriptsize
\begin{tabular}{lcccc}
\toprule
Models & \multicolumn{2}{c}{Single-label} & Multi-label & Temporal \\ \cmidrule(lr){2-3}\cmidrule(lr){4-4}\cmidrule(lr){5-5}
                       &  \begin{tabular}[c]{@{}c@{}}AID\\ \tiny(TR=20\%/50\%)\end{tabular}              & \begin{tabular}[c]{@{}c@{}}RESISC-45\\ \tiny(TR=10\%/20\%)\end{tabular}             & \begin{tabular}[c]{@{}c@{}}BEN-S2\\ \tiny(TR=10\%/100\%)\end{tabular}           & \begin{tabular}[c]{@{}c@{}}fMoW-S2\\ \tiny(TR=100\%)\end{tabular}       \\ \cmidrule(lr){2-5}
                       & OA                & OA                & mAP              & Top-1/5 Acc   \\ \midrule
GASSL \cite{ayush2021geography}                  & 93.55/95.92       & 90.86/93.06       & 79.24/87.40            & 50.69/77.99   \\
SeCo \cite{manas2021seasonal}                   & 93.47/95.99       & 89.64/92.91       & 82.62/87.81            & 51.65/77.40   \\
SatMAE \cite{cong2022satmae}                & 95.02/96.94       & 91.72/94.10                  & 86.18/89.50                 & 63.84/-              \\
RingMo \cite{sun2022ringmo}                & 96.90/98.34       & 94.25/95.67                  & -                 & -              \\
RVSA \cite{wang2022advancing}                  & 97.03/98.50       & 93.93/95.69                  & -                 & -              \\
DINO-MC \cite{wanyan2023dino}               & -                  & -                  & 84.20/88.75            & 60.16/83.49   \\
TOV \cite{tao2023tov}                   & 95.16/97.09       & 90.97/93.79                  & -                 & -              \\
SSL4EO \cite{wang2022ssl4eo}                & 91.06/94.74       & 87.60/91.27                  & 87.10/91.80                 & 51.70/76.77              \\
CMID \cite{muhtar2023cmid}                  & 96.11/97.79       & 94.05/95.53                  & -                 & -              \\
CACo \cite{mall2023change}                  & 90.88/95.05       & 88.28/91.94                  & 81.30/87.00                 & 50.72/76.31              \\
CROMA \cite{fuller2023croma}  & -                 & -                            & 88.29/-           & 63.59/-         \\ 
SatLas \cite{bastani2022satlas}                & 94.96/97.38       & 92.16/94.70                  & 82.80/88.37                 & 57.95/79.00              \\
GFM \cite{mendieta2023gfm}                  & 95.47/97.09        & 92.73/94.64                  & 86.30/-                 & -              \\
Scale-MAE \cite{reed2022scale}             & 96.44/97.58       & 92.63/95.04                  & -                 & -              \\ 
MA3E \cite{MA3E}             & - /99.04       & - /96.23                  & -                 & -              \\ 
SatMAE++ \cite{SatMAE++}  & -       & - /\textbf{97.48}                 & 85.11/-                & 63.23/-              \\ 
{SkySense} \cite{SkySense}              & {97.68/98.60}       & {94.85/96.32}                  & {88.67/92.09}                 & {64.38/87.27}              \\ \midrule
\textbf{SkySense V2}               & \textbf{98.34/99.05}       & \textbf{96.42}/97.24                & \textbf{89.13/93.78}                 & \textbf{66.65/89.32}              \\ \bottomrule
\end{tabular}
\caption{Scene classification results. - means the task is not supported or the value is unavailable in the paper.}
\vspace{-1.2em}
\label{tab:scene_classification}
\end{table}

\begin{table*}
    \begin{subtable}[t]{0.4\linewidth}
        \centering
        \setlength\tabcolsep{1.8pt}%
        \scriptsize
        \begin{tabular}{lccccc}
        \toprule
        {Models}     & {Publication}     & Dyna.-Pla. & iSAID & Potsdam & Dyna.-S2  \\ \cmidrule(lr){3-6}
                                   &                           & mIoU       & mIoU  & mF1     & mIoU      \\ \midrule
        GASSL \cite{ayush2021geography}                      & ICCV'21                   & 34.0/40.8     & 65.95 & 91.27   & 28.1/41.0 \\
        SeCo \cite{manas2021seasonal}                      & ICCV'21                   & -          & 57.20     & 89.03        & 29.4/39.8          \\
        SatMAE \cite{cong2022satmae}                    & NIPS'22                   & 32.8/39.9     & 62.97      & 90.63        & 30.1/38.7          \\
        RingMo \cite{sun2022ringmo}                    & TGRS'22                   & -          & 67.20      & 91.27        & -          \\
        RVSA \cite{wang2022advancing}                      & TGRS'22                   & 34.3/44.4     & 64.49      & -        & -          \\
        BFM \cite{cha2023billion}                        & Arxiv'23                 & -     & -      & 92.12        & -          \\
        TOV \cite{tao2023tov}                       & JSTARS'23                 & 32.1/37.8     & 66.24      & 92.03        & -          \\
        SSL4EO \cite{wang2022ssl4eo}                    & GRSM'23                   & 35.3/42.1     & 64.01      & 91.54        & 31.8/42.7          \\
        CMID \cite{muhtar2023cmid}                      & TGRS'23                   & 36.4/43.5  & 66.21      & 91.86        & -          \\
        CACo \cite{mall2023change}                      & CVPR'23                   & 35.4/42.7  & 64.32      & 91.35        & 30.2/42.5          \\
        SAMRS \cite{wang2023scaling}                 & NIPS'23                   & -     & 66.26      & 91.43        &  -         \\
        SatLas \cite{bastani2022satlas}                    & ICCV'23                   & 37.4/40.7  & 68.71      & 91.28        & 31.9/43.5          \\
        GFM \cite{mendieta2023gfm}   & ICCV'23         & 36.7/45.6        & 66.62                  & 91.85                 & -              \\
        Scale-MAE \cite{reed2022scale}                 & ICCV'23                   & 34.0/41.7     & 65.77      & 91.54        &  -         \\ 
        MA3E \cite{MA3E}                 & ECCV'24                   & -     & 64.06     & 91.50        &  -         \\ 
        {SkySense} \cite{SkySense}                  & CVPR'24                         & {39.7/46.5}           & {70.91}      & {93.99}        & {33.1/46.2}          \\ \midrule
        \textbf{SkySense V2}                   & -                         & \textbf{41.2/47.6}           & \textbf{71.87}      & \textbf{95.86}        & \textbf{35.2/47.5}          \\ \bottomrule
        \end{tabular}
        \caption{Semantic segmentation results.}
        \label{tab:seg}
        
    \end{subtable}
    \hspace{0.3em}
    \begin{subtable}[t]{0.28\linewidth}
        \centering
        \setlength\tabcolsep{1.8pt}%
        \scriptsize
        \begin{tabular}{lccc}
        \toprule
        {Models}         & Horizontal & \multicolumn{2}{c}{Oriented} \\ \cmidrule(lr){2-2}\cmidrule(lr){3-4}
                                                             & DIOR       & DIOR-R        & FAIR1M       \\ \cmidrule(lr){2-4}
                                                             & mAP$_{50}$      & mAP        & mAP          \\ \midrule
        GASSL \cite{ayush2021geography}                                         & 67.40      & 65.65         & 48.15        \\
        SatMAE \cite{cong2022satmae}                                       & 70.89      & 65.66              & 46.55             \\
        RingMo \cite{sun2022ringmo}                    & 75.90                   & -      & 46.21                          \\
        RVSA \cite{wang2022advancing}                                         & 73.22      & 71.05              & 47.04             \\
        BFM \cite{cha2023billion}                                       & -          & 73.62              & -             \\
        TOV \cite{tao2023tov}                                        & 70.16      & 66.33              & 49.62             \\
        SSL4EO \cite{wang2022ssl4eo}                                       & 64.82      & 61.23              & 49.37             \\
        CMID \cite{muhtar2023cmid}                                         & 75.11      & 66.37              & 50.58             \\
        CACo \cite{mall2023change}                                         & 66.91      & 64.10              & 47.83             \\
        SatLas \cite{bastani2022satlas}                                       & 74.10      & 67.59              & 46.19             \\
        GFM \cite{mendieta2023gfm}      & 72.84        & 67.67                  & 49.69        \\
        Scale-MAE \cite{reed2022scale}                                    & 73.81      & 66.47              & 48.31             \\ 
        MA3E \cite{MA3E}         & -      & 71.82              & -             \\ 
        {SkySense} \cite{SkySense}                                          & {78.73}           & {74.27}              & {54.57}             \\ \midrule
        {SkySense V2}                                           & \textbf{79.50}           & \textbf{75.29}              & \textbf{55.96}             \\ \bottomrule
        \end{tabular}
        \caption{Object detection results.}
        \label{tab:det}
    \end{subtable}
    \hspace{0.3em}
    \begin{subtable}[t]{0.28\linewidth}
        \centering
        \setlength\tabcolsep{1.8pt}%
        \scriptsize
        \begin{tabular}{lccc}
        \toprule
        Models  & LEVIR-CD & OSCD  & Dyna.-S2  \\ \cmidrule(lr){2-4}
                                                      & F1       & F1    & SCS       \\ \midrule
        GASSL \cite{ayush2021geography}                               & 78.19    & 46.26 & 13.6/16.7 \\
        SeCo \cite{manas2021seasonal}                                 & 90.14    & 47.67      &           13.9/16.0\\
        SatMAE \cite{cong2022satmae}                             & 87.65    & 52.76      &           14.8/16.2\\
        RingMo \cite{sun2022ringmo}                              & 91.86    & -      &           -\\
        RVSA \cite{wang2022advancing}                               & 90.86    & -      &           -\\
        SpectralGPT \cite{hong2024spectralgpt}
               & -        & 54.29      &           -\\
        MATTER \cite{akiva2022self}                               & -        & 59.37 &           -\\
        DINO-MC \cite{wanyan2023dino}                             & -        & 52.70 &           14.5/15.6\\
        SSL4EO \cite{wang2022ssl4eo}                               & 89.05    & 35.08      &           12.3/17.5\\
        CMID \cite{muhtar2023cmid}                                 & 91.72    & -      &           -\\
        CACo \cite{mall2023change}                                 & 81.04    & 52.11      &           15.3/15.8\\
        SatLas \cite{bastani2022satlas}                              & 90.62    & -      &           13.3/17.8\\
        GFM \cite{mendieta2023gfm}                                 & 91.73        & 59.82 &           -\\
        Scale-MAE \cite{reed2022scale}                           & 92.07    & -      &           -\\ 
        SkySense \cite{SkySense}                                  & {92.58}    & {60.06}      & {15.4/18.0}          \\ \midrule
        \textbf{SkySense V2}                                   & \textbf{94.83}    & \textbf{65.29}      & \textbf{16.0/20.7}          \\ \bottomrule
        \end{tabular}
        \caption{Change detection results.}
        \label{tab:change}
    \end{subtable}
    \caption{Results of semantic segmentation, object detection and change detection.}
    \vspace{-1.2em}
\end{table*}

\paragraph{Semantic Segmentation.}
In Table~\ref{tab:seg}, we report the segmentation results from four representative semantic segmentation datasets: iSAID \cite{waqas2019isaid} and Potsdam \cite{sherrah2016fully} for high-resolution optical image segmentation, and Dyna.-Pla \cite{toker2022dynamicearthnet} and Dyna.-S2 \cite{toker2022dynamicearthnet} for multi-spectral and multi-temporal image segmentation. Following previous work, we use the mean F1-score (mF1) as the evaluation metric for the Potsdam dataset. While for the other datasets, we report the mean intersection over union (mIoU). All segmentation experiments employ the UperNet \cite{xiao2018unified} as the decoder head. As illustrated in Table~\ref{tab:seg}, our SkySense V2 achieve higher performance than recent RSFMs across all four datasets. Specifically, compared to the previous SOTA method SkySense \cite{SkySense}, SkySense V2 achieves 1.5\% higher performance on average. 

\paragraph{Horizontal \& Oriented Object Detection.}
We utilize the DIOR dataset \cite{li2020object} to evaluate SkySense V2's performance in horizontal object detection, while utilizing the DIOR-R \cite{cheng2022anchor} and FAIR1M \cite{sun2022fair1m} datasets for oriented object detection. These datasets only consist of RGB RS imagery. Consistent with previous research \cite{SkySense}, we adopt Faster R-CNN  \cite{ren2015faster} and Oriented R-CNN \cite{li2022oriented} as the basic detectors for horizontal and oriented object detection, respectively. As reported in Table~\ref{tab:det}, our SkySense V2 outperforms previous RSFMs by a notable margin. Specifically, SkySense V2 surpass the previous SOTA, SkySense \cite{SkySense}, by an average of 1.1\% mAP. 

\paragraph{Change Detection.}
For change detection, we conducted experiments on LEVIR-CD \cite{chen2020spatial}, OSCD \cite{daudt2018urban} and Dyna.-S2 \cite{toker2022dynamicearthnet} datasets. Since LEVIR-CD and OSCD datasets only contain binary change detection results, we use the basic change detector in \cite{chen2021remote} and report the F1 score as the evaluation metric. While Dyna.-S2 focuses on semantic change detection, we use UperNet \cite{xiao2018unified} as the segmentation network and calculate the semantic change detection score (SCS) on validation and test set following \cite{SkySense}. As presented in Table~\ref{tab:change}, SkySense V2 consistently achieves higher performance than the previous SOTA method SkySense on all 3 datasets, and surpasses SkySense by 2.7\% on average. Particularly, on the OSCD dataset, our SkySense V2 remarkably outperform SkySense by 5.2\% on F1-score. 

\subsection{Multi-modal Tasks}
In this section, we evaluate SkySense V2 on two representative multi-modal tasks, \ie, Multi-modal Segmentation and Multi-modal Scene Classification, to demonstrate its generalization capability across data from various modalities.

\begin{table}
    \centering
    \setlength\tabcolsep{2.5pt}%
    \scriptsize
    \begin{tabular}{llcc}
        \toprule
        \textbf{\begin{tabular}[c]{@{}l@{}}Task \& Dataset\end{tabular}} & \textbf{Data Source} & \textbf{SkySense}\cite{SkySense} & \textbf{SkySense V2} \\ 
        \midrule
        \multirow{4}{*}{\begin{tabular}[c]{@{}l@{}}(a) Multi-modal Seg:\\ Dyna.-MM\end{tabular}} 
        & \texttt{(i)} Planet.                & 46.5 &  \textbf{47.6}\\ 
        & \texttt{(ii)} S2                   & 46.2 &  \textbf{47.5} \\ 
        & \texttt{(iii)} Planet. + S2          & 47.3 &  \textbf{48.7}\\ 
        & \texttt{(iv)} Planet. + S2 + S1      & 47.7 &   \textbf{48.9} \\ 
        \midrule
        \multirow{4}{*}{\begin{tabular}[c]{@{}l@{}}(b) Multi-modal Seg: \\ PASTIS-MM\end{tabular}} 
        & \texttt{(i)} S2                 & 73.5 &  \textbf{75.0}\\ 
        & \texttt{(ii)} S2-Ts                 & 84.6 &  \textbf{85.5}\\ 
        & \texttt{(iii)} S2-Ts + S1-Ts        & 84.8 &  \textbf{85.6}\\ 
        & \texttt{(iv)} S2-Ts + GEP           & 85.8 &  \textbf{86.7}\\ 
        \midrule
        \multirow{2}{*}{\begin{tabular}[c]{@{}l@{}}(c) Multi-modal Cls: \\ BEN-MM\end{tabular}} 
        & \texttt{(i)} S1                     & 86.2 & \textbf{86.5} \\ 
        & \texttt{(ii)} S2 + S1               & 92.2 & \textbf{93.8} \\ 
        \bottomrule
    \end{tabular}
    \caption{Fine-tuning results on multi-modal tasks.}
    \vspace{-1em}
    \label{tab:mm_tasks}
\end{table}

\paragraph{Multi-modal Segmentation.}
In Table~\ref{tab:mm_tasks} (a) and (b), we present the performance results of Dyna.-MM \cite{toker2022dynamicearthnet} and PASTIS-MM \cite{garnot2022multi}, respectively. Dyna.-MM contains high-resolution optical imagery from PlanetFusion (Planet.), multi-spectral imagery from Sentinel-2 (S2), and SAR imagery from Sentinel-1 (S1). For the experiments conducted on Dyna.-MM, we employ a basic UperNet \cite{xiao2018unified} as the decoder head and utilize mIoU as the evaluation metric. As depicted in Table~\ref{tab:mm_tasks} (a), SkySense V2 consistently outperforms the SkySense across various modalities, demonstrating a stronger representation ability. Furthermore, integrating multiple modalities yields higher performance compared to using a single modality. This results confirms that the proposed SkySense V2 can effectively extract representations from diverse modalities using a unified backbone.

\par PASTIS-MM is a crop mapping dataset that contains high-resolution optical imagery from Google Earth Pro (GEP), multi-temporal multi-spectral imagery from Sentinel-2 (S2-Ts), and multi-spectral SAR imagery from Sentinel-1 (S1-Ts). We employ a straightforward FCN head to decode the semantic segmentation results and report the overall accuracy as the evaluation metric in Table~\ref{tab:mm_tasks} (b). The results demonstrate that SkySense V2 outperforms the previous SOTA method SkySense by an average of 1.0\%. Additionally, it is worth noting that S2-Ts can remarkably improve the segmentation accuracy compared to S2, underscoring the importance of temporal information for crop mapping.

\vspace{-10pt}

\paragraph{Multi-modal Scene Classification.}
We conduct multi-modal scene classification experiments on the BEN-MM dataset \cite{Sumbul2021bigearthnet}. As is shown in Table~\ref{tab:mm_tasks} (c), the proposed SkySense V2 surpasses SkySense, demonstrating a stronger representation ability. Furthermore, by integrating S1 and S2 imagery, SkySense V2 achieves higher improvements, highlighting its enhanced ability to extract superior representations from different modalities compared to SkySense.

\subsection{Ablation and Discussion}

\begin{figure}[!htp]
    \centering  
   \begin{subfigure}{0.49\linewidth} 
      \centering   
      \includegraphics[width=1\linewidth]{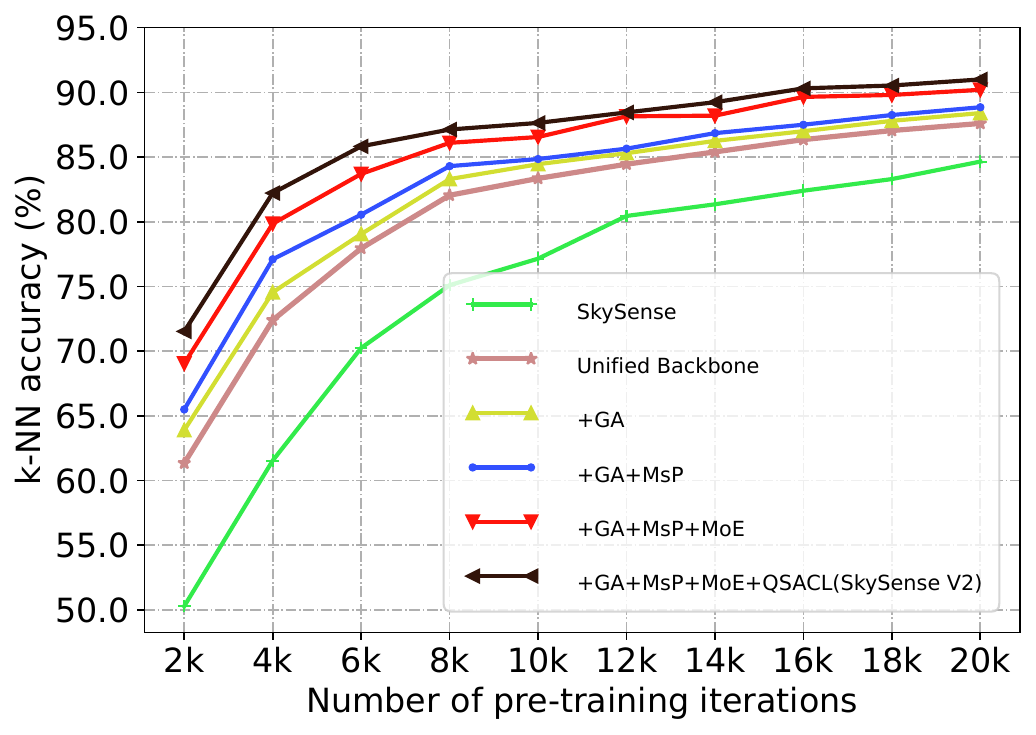}
        \caption{AID dataset}
        \label{fig:ablation_knn_AID}
    \end{subfigure} 
    \begin{subfigure}{0.49\linewidth} 
      \centering   
      \includegraphics[width=1\linewidth]{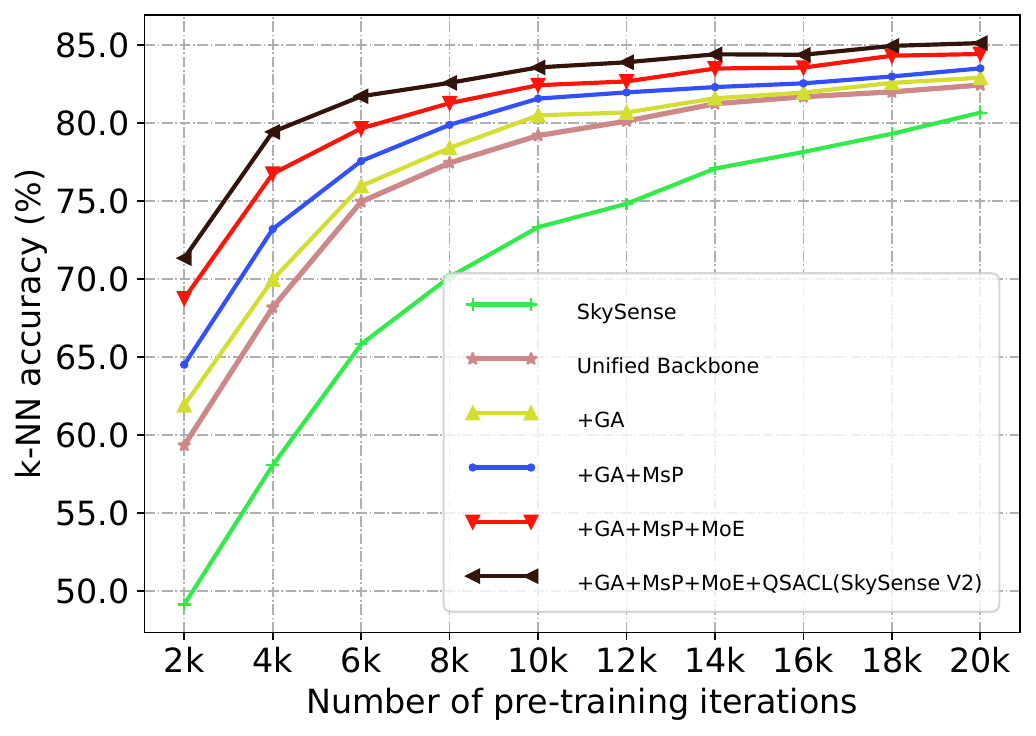}
        \caption{RESISC-45 dataset}
        \label{fig:ablation_knn_nwpu}
    \end{subfigure} 
    \caption{Ablation study using k-NN classification was conducted on the AID and RESISC-45 datasets. In this context, the abbreviations have the following meanings: GA refers to replacing window-based self-attention with global self-attention in the last two stages; MsP involves the addition of modality-specific prompt tokens; MoE indicates scaling up the model through a mixture of experts approach; and QSACL means adding additional query-based semantic aggregation contrastive learning.
    }
    \label{fig:abation_knn}
\end{figure}

\paragraph{Ablation Study of Components.} To understand the contributions of each component, we conduct a k-nearest neighbors (k-NN) evaluation on the AID and RESISC-45 datasets. Figure \ref{fig:abation_knn} illustrates the performance of different component combinations across various pre-training iterations. Notably, the unified backbone design significantly accelerates representation learning. This is primarily due to shared parameters across different modalities, which allow gradients to aggregate, thereby speeding up convergence. Moreover, the unified design allows the backbone to be trained with data from different modalities, thereby enhancing the model's generalization capabilities (see Appendix F.3 for more details). The results also indicate that global self-attention complements window-based self-attention, facilitating improved representations. Additionally, modality-specific prompt tokens and MoE scaling enhance representation learning by increasing feature diversity and model capacity, respectively. QSACL enhances model performance by improving the semantic precision of features in contrastive learning.
\begin{figure}[!htp]
    \centering  
   \begin{subfigure}{0.49\linewidth} 
      \centering   
      \includegraphics[width=1\linewidth]{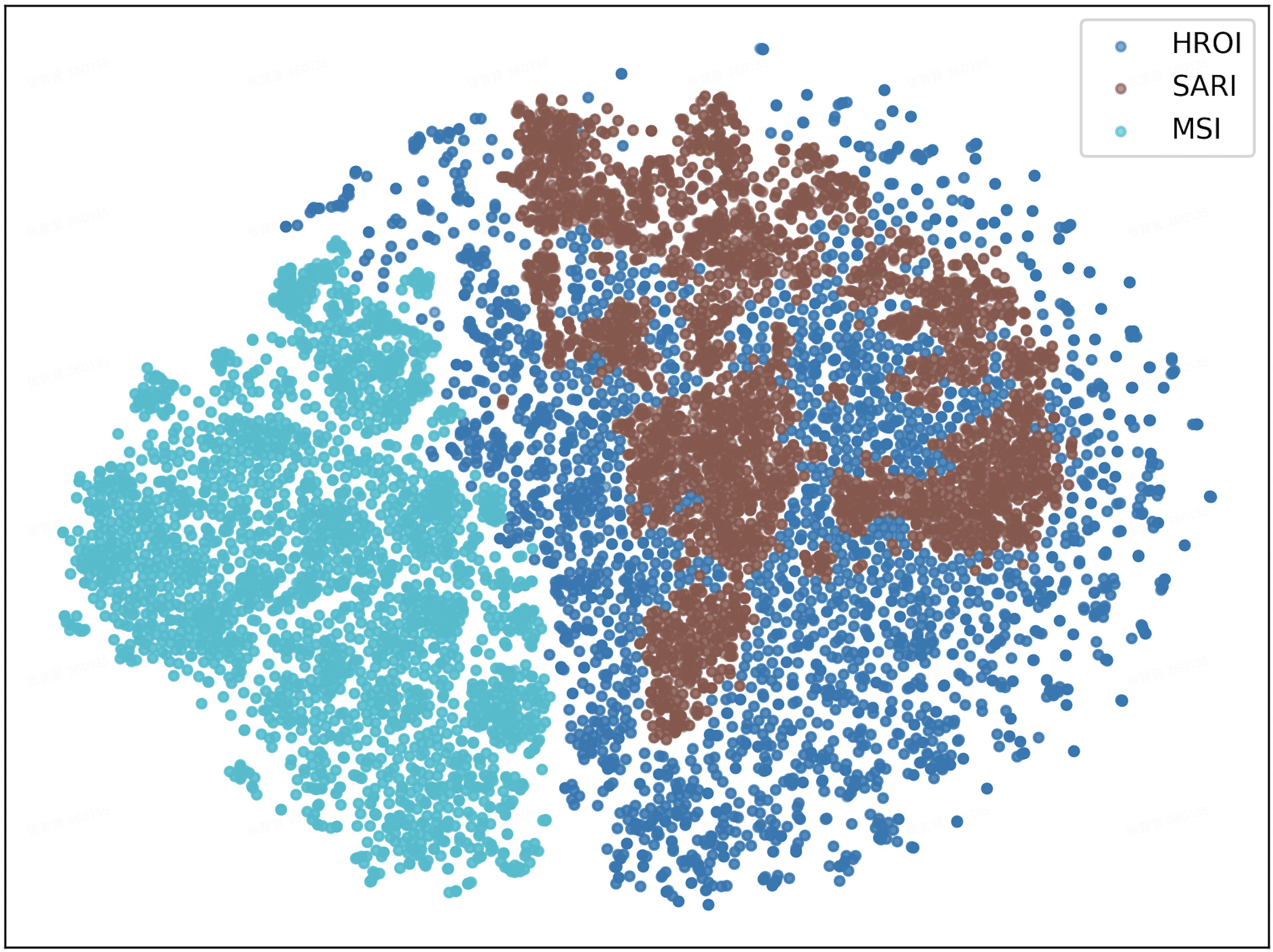}
        \caption{w/o modality-specific prompt}
        \label{fig:without_msp}
    \end{subfigure} 
    \begin{subfigure}{0.49\linewidth} 
      \centering   
      \includegraphics[width=1\linewidth]{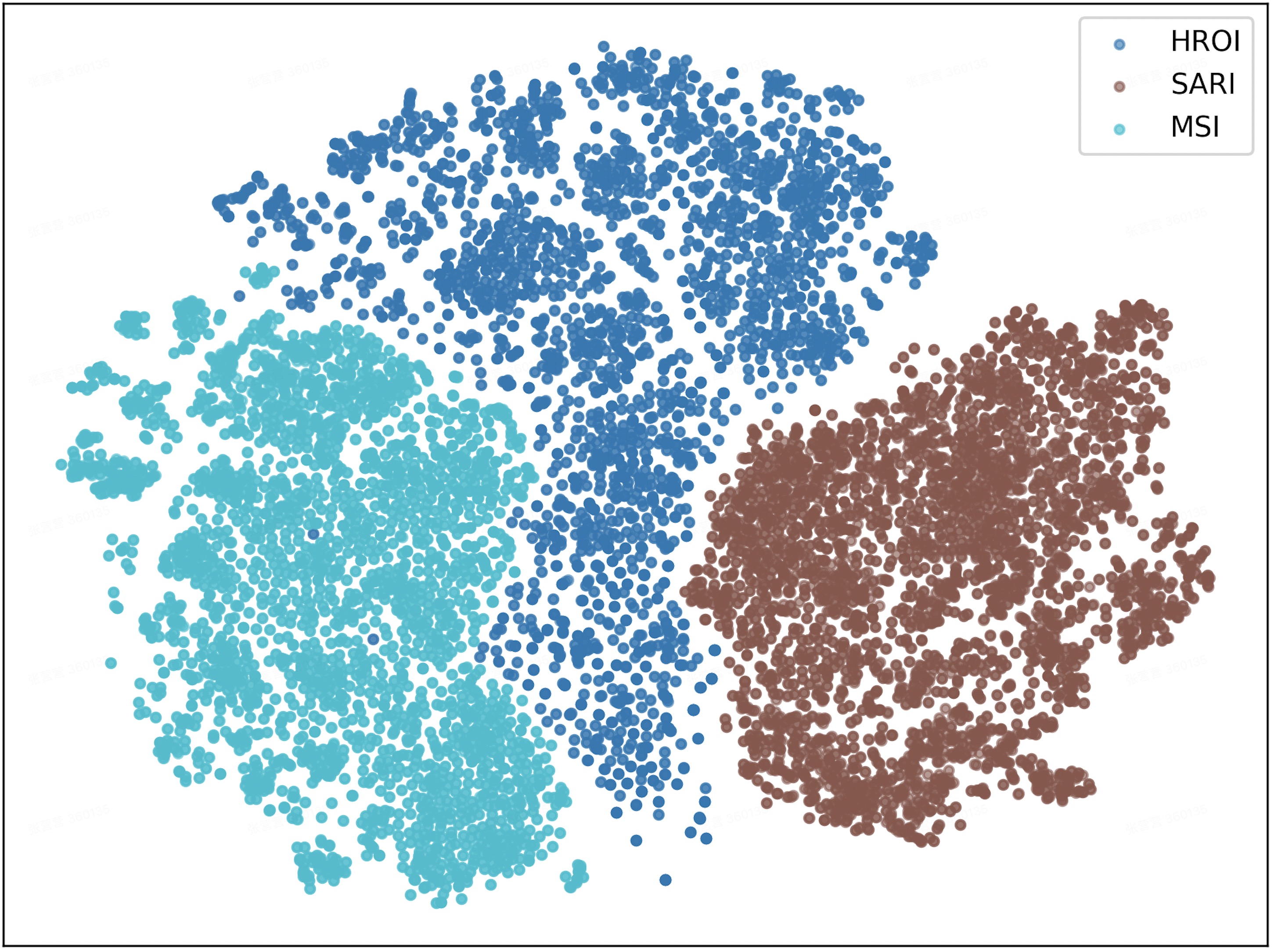}
        \caption{w/ modality-specific prompt}
        \label{fig:with_msp}
    \end{subfigure} 
    \caption{t-SNE visualization of the output feature representation from the final block of the unified transformer backbone. HROI, SARI, and MSI represent the features from high-resolution optical imagery, synthetic aperture radar imagery, and multi-spectral imagery, respectively.}
    \label{fig:tsne_vis}
\end{figure}

\paragraph{How Do Modality-Specific Prompt Tokens Facilitate Pre-training?} To investigate the impact of modality-specific prompt (MsP) tokens on the multi-modal pre-training process within a unified backbone, we employed t-SNE \cite{t-SNE} for visualizing the features corresponding to different modalities. These features are extracted from the last block of the unified transformer backbone, and the comparative results are presented in Figure \ref{fig:tsne_vis}. Without the use of MsP, the features of high-resolution optical imagery (HROI) and synthetic aperture radar imagery (SARI) appear closely clustered, leading to overlapping feature distributions. However, with the inclusion of MsP, the features of HROI and SARI become distinctly separable. These visualization outcomes clearly demonstrate that MsP effectively enhances feature diversity and imparts modality-specific characteristics.

\begin{figure}[H]
	\centering
		\includegraphics[width=0.48\textwidth]{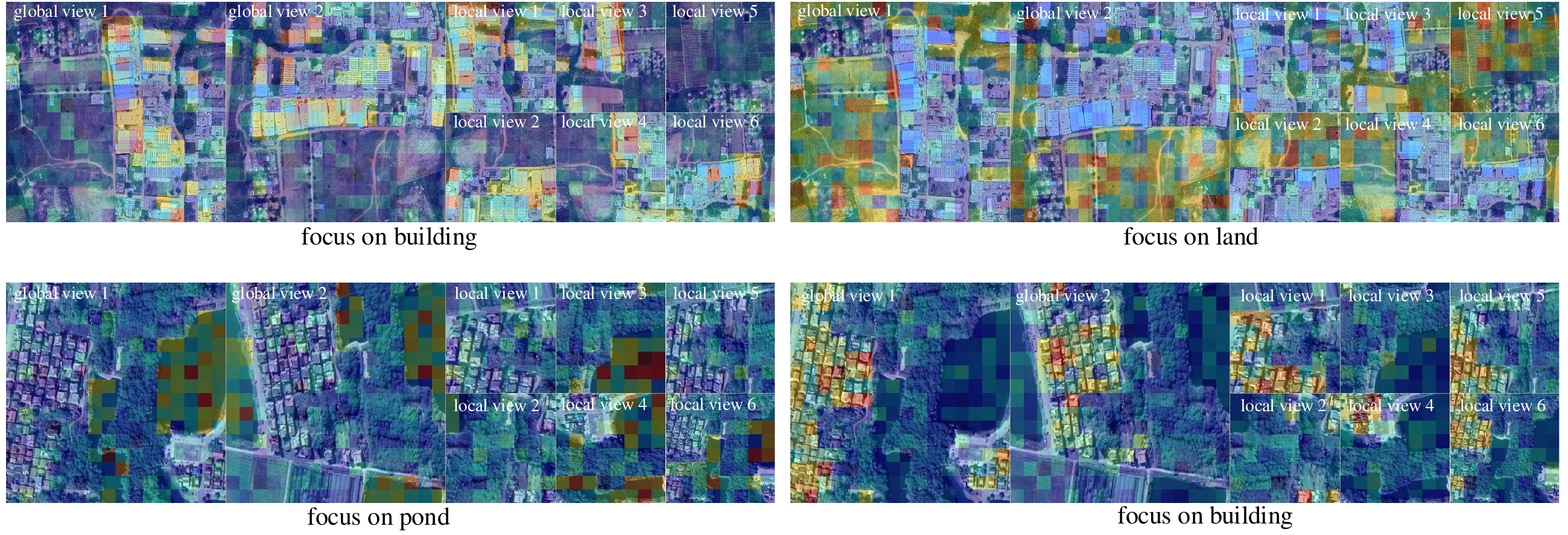}
	    \caption{Visualization of cross-attention weights for features patch corresponding to different queries in QSACL. Queries can effectively aggregate features with specific semantics. }
	\label{qsacl_attn}

\end{figure}

\paragraph{What Types of Features Does Query-based Semantic Aggregation Contrastive Learning Capture?} To provide a clearer understanding of our proposed QSACL, we visualize the attention weights assigned to different features during their interaction with certain query, as illustrated in Figure \ref{qsacl_attn}. During pre-training, we use two global augmentation crops and six local augmentation crops. Here, we present two distinct queries, each focusing on different semantic features of image patches. For example, one query focuses on features with building semantics (left part of the first row in Figure \ref{qsacl_attn}), while another focuses on features with land semantics (right part of the first row in Figure \ref{qsacl_attn}). The attention weights of different queries aggregate features from each patch into distinct feature representations. Subsequently, we perform contrastive learning between aggregated features from the same query across teacher and student models. This approach significantly enhances semantic accuracy compared to traditional contrastive learning applied to entire image features.


\section{Conclusion}

In this paper, we introduce SkySense V2, a MM-RSFM that utilizes a unified backbone to accommodate various modalities. This unified approach enhances parameter utilization efficiency and improves the model's generalization ability. SkySense V2 is pre-trained with an innovative QSACL strategy, specifically designed to leverage the unique characteristics of RS images. As a result, SkySense V2 significantly boosts performance while maintaining the multi-modal flexibility benefits of its predecessor, SkySense. Looking ahead, our future work will focus on integrating the language modality and incorporating a geographical knowledge graph to develop a more powerful and versatile MM-RSFM.

\appendix

\section{Pre-training Module \& Loss}

\subsection{Multi-Granularity Contrastive Learning} 
We implement the multi-granularity contrastive learning proposed in SkySense\cite{SkySense} for self-supervised learning across multiple modalities and spatial granularities. Given the input set $\left\{x_{HR}, x_{MS}, x_{SAR}\right\}$, two separate collections of augmented views, denoted as $\left\{u_{i}\right\}$ and $\left\{v_{i}\right\}$, are generated through random augmentations, where where $i\in \{HR, MS, SAR\}$. These views $u_i$ and $v_i$ are then input into the student and teacher branches respectively. In the student branch, let $\mathcal{T}_i$ represent the tokenizer for each modality and $\mathcal{U}$ the unified transformer backbone of SkySense V2. The weights for the teacher branch are calculated as the exponential moving average (EMA) of the student branch weights: $\mathcal{T}_i^{\prime} = EMA(\mathcal{T}_i)$ , $\mathcal{U}^{\prime} = EMA(\mathcal{U})$s. This procedure yields spatial features as described in Equation~\ref{backbone_ema}:
\begin{equation}
F_i=\mathcal{U}(\mathcal{T}_i\left(u_i\right)), F_i^{\prime}=\mathcal{U}^{\prime}(\mathcal{T}_i^{\prime}\left(v_i\right)) ~~i \in\{HR, MS, SAR\}.
\label{backbone_ema}
\end{equation}
By applying multi-modal temporal fusion and geo-context integration \cite{SkySense} to $F_i$ and $F_i^{\prime}$, we obtain the final features $F_{\mathbf{fus}}$ and $F_{\mathbf{fus}}^{\prime}$. We then initiate pixel-level, object-level, and image-level contrastive learning to progressively acquire coarse-to-fine spatial features for various tasks.
\paragraph{Pixel-level Loss.}Each temporal slice of spatial feature $F_i$ can be viewed as a pixel-level feature $F_i^{\mathbf{pix}} \in \mathbb{R}^{N_S \times d}$. The pixel-level contrastive learning loss, denoted as $\mathcal{L}_{\mathbf{pix}}$ is calculated by averaging all $\mathcal{L}_{CL}$ over both spatial ($s$) and temporal ($t$) dimensions, as described in Equation~\ref{equ:pixel}. Here, $f_i^{\mathbf{pix}} \in \mathbb{R}^{d}$ represents a feature vector from $F_i^{\mathbf{pix}}$ in specific location, and $f_i^{\mathbf{pix}\prime}$ is its correspondence at the same geo-location. $\mathcal{L}_{CL}$ denotes the learning loss~\cite{caron2021emerging} between $f_i^{\mathbf{pix}}$ and $f_i^{\mathbf{pix}\prime}$:
\begin{equation}
\begin{aligned}
\mathcal{L}_{\mathbf{pix}}(F_{i}, F_{i}^{\prime}) = \frac{1}{N_S T_i}\sum_s \sum_t \mathcal{L}_{CL}(f_i^{\mathbf{pix}}, f_i^{\mathbf{pix}\prime}).
\end{aligned}
\label{equ:pixel}
\end{equation}

\paragraph{Object-level Loss.} The object-level features $F_i^{\mathbf{obj}} \in \mathbb{R}^{N_C \times d}$ are generated from unsupervised clustering on pixel-level feature vectors $f_i^{\mathbf{pix}}$ in a single RSI, where $N_C$ is the number of clusters. For clustering, we employ the Sinkhorn-Knopp algorithm~\cite{caron2020unsupervised}, as used in \cite{SkySense}. Each cluster center, denoted as $f_i^{\mathbf{obj}} \in \mathbb{R}^{d}$ serves as a generalized representation for a collection of $f_i^{\mathbf{pix}}$. This cluster center typically corresponds to a specific ground object or semantic concept. We calculate the object-level contrastive learning loss as follows:
\begin{equation}
\begin{aligned}
\mathcal{L}_{\mathbf{obj}}(F_{i}, F_{i}^{\prime}) = \frac{1}{N_C T_i}\sum_s \sum_t \mathcal{L}_{CL}(f_i^{\mathbf{obj}}, f_i^{\mathbf{obj}\prime}).
\end{aligned}
\label{equ:object}
\end{equation}

\paragraph{Image-level Loss.} The image-level feature $F_i^{\mathbf{img}} \in \mathbb{R}^{d}$ is simply an average pooling result from $F_i^{\mathbf{pix}}$. The image-level contrastive learning loss is defined as follows:
\begin{equation}
\begin{aligned}
\mathcal{L}_{\mathbf{img}}(F_{i}, F_{i}^{\prime}) = \frac{1}{T_i}\sum_t \mathcal{L}_{CL}(F_i^{\mathbf{img}}, F_i^{\mathbf{img}\prime}).
\end{aligned}
\label{equ:image}
\end{equation}

Finally, the fine-grained contrastive learning loss $\mathcal{L}_{FGCL}$ is the sum of pixel-, object- and image-level contrastive learning losses, as described in Equation~\ref{equ:fgcl}. Subsequently, we develop multi-modal loss $\mathcal{L}_{MGCL}$ as shown in Equation~\ref{deqn_contras_modal}. The multi-granularity concept is reflected in two main dimensions: spatial and modal. From a spatial perspective, contrastive learning is executed at the pixel, object, and image levels, enabling representation learning that comprehensively captures different spatial dimensions. From a modal perspective, we perform contrastive learning on both the features of individual modalities, denoted as $F_i$, the fused multi-modal features, represented as, $F_{\mathbf{fus}}$:
\begin{equation}
\begin{aligned}
\mathcal{L}_{FGCL}(F_{i}, F_{i}^{\prime}) = \sum_{n \in \left\{\mathbf{pix}, \mathbf{obj}, \mathbf{img}\right\}} \mathcal{L}_{n}(F_{i}^{}, F_{i}^{\prime}),
\end{aligned}
\label{equ:fgcl}
\end{equation}

\begin{equation}
\begin{aligned}
\mathcal{L}_{MGCL} & = \sum_{i \in \left\{HR, MS, SAR\right\}} \mathcal{L}_{FGCL}(F_{i}, F_{i}^{\prime}) \\
& + \mathcal{L}_{FGCL}(F_{\mathbf{fus}}, F_{\mathbf{fus}}^{\prime}).
\end{aligned}
\label{deqn_contras_modal}
\end{equation}


\subsection{Dense Image-Text Alignment} 
In addition to the $\mathcal{L}_{MGCL}$ and $\mathcal{L}_{QSACL}$ losses, we introduce an auxiliary supervision strategy using OpenStreetMap (OSM)\footnote{https://www.openstreetmap.org/} to enhance dense interpretation capabilities. OSM is an open-source, global-scale database that provides pixel-level land-cover and land-use categories. For multi-modal input imagery, we first collect the corresponding pixel-level OSM labels. Each pixel's class name is converted into a text representation using the CLIP \cite{radford2021learning} text encoder, and its visual representation is aligned with this text representation. Our experiments demonstrate that this dense image-text alignment encourages SkySense V2 to learn dense and semantic-aware representations. 
\par Specifically, assuming the category set of OSM includes $K$ classes, we first encode all class names to text representations $F^{text}\in\mathbb{R}^{K\times D}$ with the CLIP text encoder, where $D$ denotes the number of feature dimensions. Given a vision feature $F\in \mathbb{R}^{N\times D}$ extracted by the SkySense V2 backbone, we maximize the similarity between each pixel's vision feature and its corresponding text feature while minimizing the similarity with non-matching text features. The dense image-text alignment loss $\mathcal{L}_{ITA}$ is then formulated as 
\begin{equation}
\begin{aligned}
\mathcal{L}_{ITA} = -\frac{1}{n}\log(\sum_{i\in n}\frac{\exp(F_i*F^{text}_j/\tau)}{\sum_{k=1}^{K}\exp(F_i*F^{text}_k/\tau)}),
\end{aligned}
\label{eq:loss_ita}
\end{equation}
where $j$ denotes the label index of the $i$-th vision feature, and $\tau$ is a temperature parameter that controls the smoothness of the logits. By aligning the vision and text representations for every pixel as described in Eq.~\ref{eq:loss_ita}, SkySense V2 generates a more fine-grained interpretation of the input imagery.

\subsection{Unsupervised Geo-Context Prototype Learning} 
Different regions are characterized distinct geographic landscapes~\cite{huang2022toward,hu2020unsupervised} influenced by variations in culture, topography, and climate. SkySense~\cite{SkySense} has demonstrated that this regional geo-context benefits the interpretation of remote sensing imagery~\cite{liu2023seeing,chen2019collaborative,guo2022isdnet,huang2022toward}. Following the approach of SkySense \cite{SkySense}, we employ unsupervised geo-context prototype learning (GCPL) to group similar $F_{\mathbf{fus}}^{\text{mm}}$. And these features are integrated as implicit geo-knowledge over a wide geo-spatial range to augment original feature during pre-training. Specifically, we divide the globe into $N_{R}$ regions and initialize a region-specific prototype set $\mathcal{P} \in \mathbb{R}^{N_{R} \times N_{p} \times d}$. Each prototype is learned based on $F_{\mathbf{fus}}^{\text{mm}}$. We leverage the geo-location of the RSI to retrieve the regional subset $\mathcal{P}_r \in \mathbb{R}^{N_{p} \times d}$  from $\mathcal{P}$. Then, we calculate the cosine similarity matrix $\mathbf{M} \in \mathbb{R}^{N_{S} \times N_{p}}$ between $F_{\mathbf{fus}}^{\text{mm}}$ and $\mathcal{P}_r$:
\begin{align}
\label{deqn_cosm}
\mathbf{M} = \frac{F_{\mathbf{fus}}^{\text{mm}} \cdot \mathcal{P}_r^{\text{T}}}{\Vert F_{\mathbf{fus}}^{\text{mm}} \Vert \Vert \mathcal{P}_r \Vert}.
\end{align}

The Sinkhorn-Knopp (SK) algorithm \cite{caron2020unsupervised} on $\mathbf{M}$ is utilized to find the optimal assignment matrix $\mathbf{S} \in \mathbb{R}^{N_{S} \times N_{p}}$ between $F_{\mathbf{fus}}^{\text{mm}}$ and the prototypes. The SK algorithm incorporates a uniform distribution constraint to circumvent trivial solutions while striving to achieve the highest similarity possible. Subsequently, we utilize $\mathbf{S}$ to generate an updated value for current sample's corresponding $\mathcal{P}_r$, denoted as $\overline{\mathcal{P}_r}$. This process is detailed as follows:
\begin{align}
\label{deqn_ex2}
\overline{\mathcal{P}_r} = \mathbf{S}^{\text{T}}F_{\mathbf{fus}}^{\text{mm}}.
\end{align}
Afterwards, we update $\mathcal{P}_r$ through EMA~\cite{he2020momentum} as in Equation~\ref{deqn_ex3}, where $m \in [0, 1)$ is a momentum coefficient.
\begin{align}
\label{deqn_ex3}
\mathcal{P}_{r}\leftarrow m \mathcal{P}_r + (1-m) \overline{\mathcal{P}_r}.
\end{align}
Each $\mathcal{P}_{r}$ is updated during pre-training and serves as a fixed geo-context for downstream tasks. GCPL is applied exclusively to the student branch, extracting generalized region-aware representations from numerous RSI within a consistent region. This provides complementary information to enhance the features of individual RSI.

\section{Downstream Usage of SkySense V2}

\begin{figure}
	\centering
		\includegraphics[scale=.34]{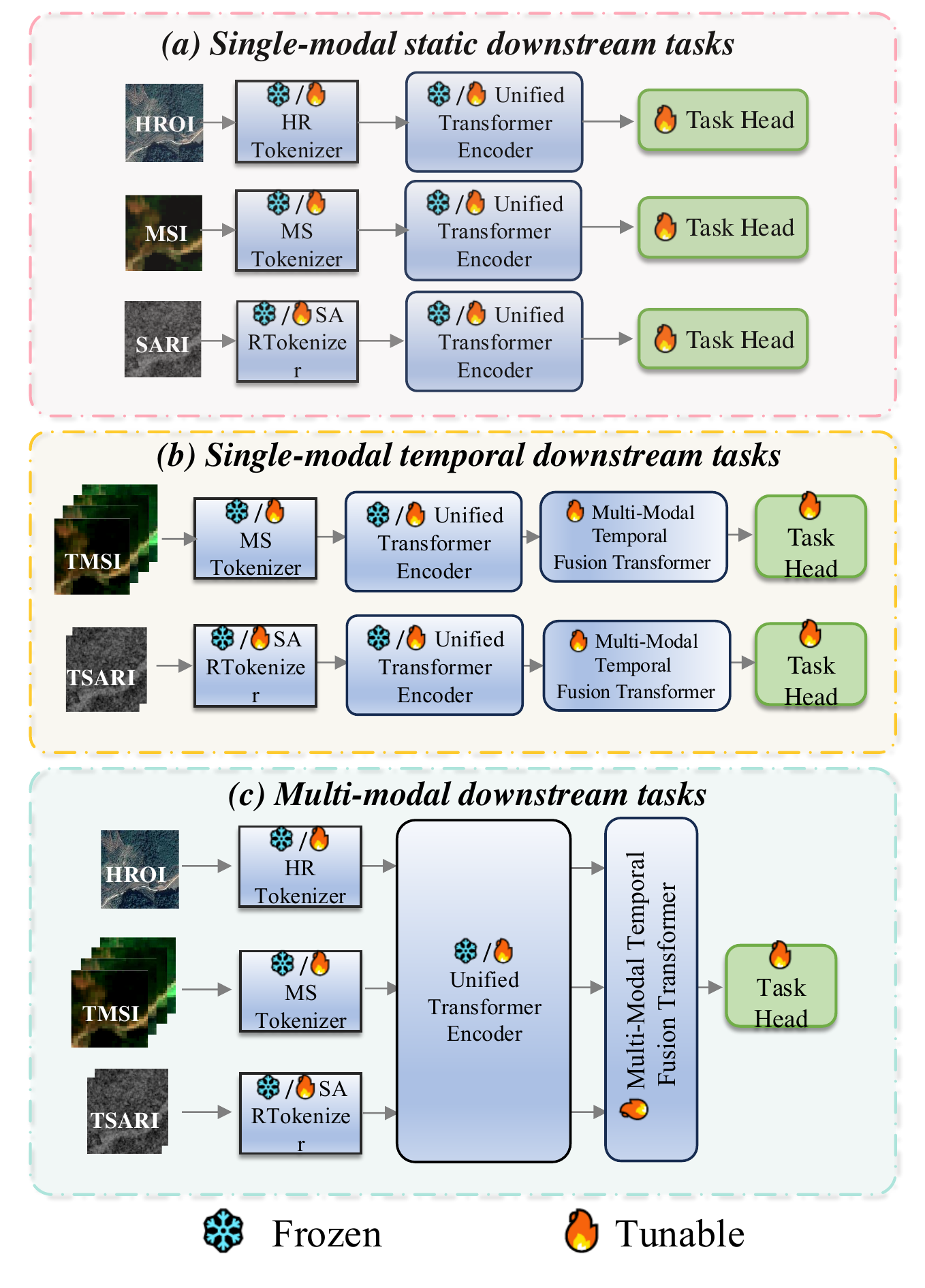}
	    \caption{Overview of Downstream Usage of SkySense V2. Each pre-trained module can be utilized independently or in combination, with options to freeze or fine-tune the selected modules based on the specific downstream task requirements.}
	\label{fig:downstream_usage}
    \vspace{-10pt}
\end{figure}

After pre-training, we utilize the parameters from the teacher branch for downstream tasks, as shown in Figure~\ref{fig:downstream_usage}. Each pre-trained module can be used independently or in combination with others, with the selected modules either frozen or fine-tuned. For single-modal static downstream tasks, we retain the unified transformer backbone and activate the specific tokenizer. Additionally, we add a task-specific head tailored to the particular task. In single-modal temporal downstream tasks, we incorporate the pre-trained fusion transformer to process time series feature data from a single modality. This fusion transformer integrates temporal information, enabling the model to capture dynamic patterns and trends over time, which are crucial for applications such as crop identification or change detection. For multi-modal downstream tasks, the fusion transformer is employed to integrate features from different modalities. This integration addresses both modality-specific and temporal aspects, allowing the model to leverage complementary information from various data sources. By fusing multi-modal data, SkySense V2 enhances its ability to perform complex tasks that require the synthesis of diverse information. This flexibility ensures that SkySense V2 can be effectively applied to a wide range of downstream applications, maintaining high performance while adapting to varying task demands.

\section{Pre-training Implementation Details}
SkySense V2 is pre-trained using a batch size of 1024, distributed across 128 H20 GPUs. The model undergoes a total of 600k iterations, utilizing the AdamW optimizer~\cite{loshchilov2018fixing} with $\beta_1=0.9, \beta_2=0.999$. The learning rate is initially set to $2 \times 10 ^ {-4}$ and decays to $1 \times 10 ^ {-6}$ following a cosine annealing schedule \cite{SGDRSG}. Similarly, the weight decay follows a cosine schedule, starting at 0.04 and increasing to 0.2. Additionally, to maintain stable training, the gradient is clipped at an $L_2$ norm of 3.0 for all parameters. The momentum in EMA updating for teacher network is initialized as 0.996 and decay to 1.0 with cosine schedule. The loss weights for loss $\mathcal{L}_{MGCL}, \mathcal{L}_{QSACL}, \mathcal{L}_{ITA}$ are set as: $\lambda_1=1.0, \lambda_2=1.0, \lambda_3=1.0$. The weight of MoE auxiliary loss is set to 0.01. The number of queries of QSACL is set to 16.  The whole pre-training progress takes 44500 H20 GPU hours, and its computational complexity is 8109.52 GFLOPs.

For high-resolution optical imagery (HROI), we apply augmentations including Gaussian blur, solarization~\cite{grill2020bootstrap}, random color jitter, random flips, and random rotations. In terms of multi-spectral imagery (MSI) and synthetic aperture radar imagery (SARI) time series, we randomly select a fixed-sized sequence (20 for MSI and 10 for SARI) from the original one and perform random disturbances on the RSI acquisition date. We follow the global and local multi-view cropping strategy in ~\cite{caron2020unsupervised, SkySense}, with 2 global views and 6 local views being used respectively.

Following SkySense\cite{SkySense}, the multi-modal temporal fusion transformer module contains 24 basic transformer encoder layers. Additionally, a single basic transformer decoder layer is employed for query-based semantic aggregation contrastive learning. For GCPL, the globe is segmented into 4096 regions, each covering an area of roughly 4294 square kilometers and consisting of 100 prototypes.

\section{Downstream Tasks Training Implementation Details}
\subsection{Semantic Segmentation} 
\label{semantic_segmentation}
Semantic segmentation is widely used in remote sensing to automatically extract land use classes and ground instances. Considering factors such as spatial resolution, spectrum and number of categories, we select four popular datasets for the semantic segmentation task: DynamicEarthNet-PlanetFusion (Dyna.-Pla.) \cite{toker2022dynamicearthnet}, iSAID \cite{waqas2019isaid}, Potsdam \cite{sherrah2016fully}, and DynamicEarthNet-Sentinel2 (Dyna.-S2). We employ the UperNet \cite{xiao2018unified} as the unified segmentation head,implemented based on the MMSegmentation\footnote{\url{https://github.com/open-mmlab/mmsegmentation}}, in line with the approaches of \cite{sun2022ringmo, cha2023billion, wang2022advancing}. Detailed fine-tuning settings are provided in Table~\ref{seg_setting}.

\begin{table}
        \centering
        \setlength\tabcolsep{2pt}%
        \small
        \begin{tabular}{l|cccc}
        \hline
        Dataset & Dyna.-Pla. & iSAID & Potsdam & Dyna.-S2\\ 
        \hline
        \begin{tabular}[c]{@{}c@{}}Activated \\ modality \end{tabular} & HR & HR & HR & MS \\
        Optimizer  & AdamW & AdamW & AdamW & AdamW    \\
        Input size & 1024$\times$1024  & 896$\times$896 & 512$\times$512   & 256$\times$256 \\         
        Input channel & RGBNIR & RGB & NIRRG  & \begin{tabular}[c]{@{}c@{}}B02-08, B8A, \\ B11-12\end{tabular} \\ 
        Base lr.  & 1e-4 & 1e-4& 1e-4& 1e-4 \\
        Lr. scheduler & poly & poly & poly & poly  \\
        Weight decay & 0.01 & 0.01 & 0.01 & 0.01  \\
        \begin{tabular}[c]{@{}c@{}}Layer-wise \\ lr decay\end{tabular}& 0.8   & 0.8    & 0.8   & 0.8  \\
        Max iters. & 80k & 80k  & 80k  & 80k    \\  
        Warmup  &linear &linear &linear &linear    \\
        Warmup iters. & 1.5k & 1.5k  & 1.5k  & 1.5k  \\
        Warmup ratio & 1e-6 & 1e-6 & 1e-6 & 1e-6 \\
        Drop path rate  &0.2 &0.2 &0.2 &0.2 \\
        Augmentations & & & & \\
        RandomScaling & &\checkmark &\checkmark &\\
        RandomCrop &\checkmark &\checkmark &\checkmark &\checkmark \\
        RandomFlip &\checkmark &\checkmark &\checkmark &\checkmark \\
        \hline
        \end{tabular}
    \caption{The finetuning setting in single-modal semantical segmetation tasks. The minimum and maximum values for random scaling are 0.5 and 2.0, respectively, and the probability of a random flip is 0.5.}
    \label{seg_setting}
\end{table}

\subsection{Horizontal \& Oriented Objection Detection} 
Remote sensing images encompass a diverse array of objects, including buildings, vehicles, bridges and so on. These objects are densely distributed and vary widely in size, scale, and orientation, making their detection and identification a challenging task \cite{wen2023comprehensive}. To evaluate the effectiveness of RSFMs in oriented object detection, we use the DIOR-R and FAIR1M datasets and implement the Oriented RCNN \cite{li2022oriented} as the detection algorithm, in line with prior studies \cite{SkySense, sun2022ringmo,wang2022advancing,cha2023billion}. For assessing the horizontal object detection capabilities of SkySense V2, we utilize the DIOR dataset. Following the methodology of \cite{SkySense,sun2022ringmo}, we employ the Faster RCNN \cite{ren2015faster} as the detector. Additional details are provided in Table~\ref{det_setting}.

\begin{table}
        \centering
        \setlength\tabcolsep{2pt}%
        \small
        \begin{tabular}{l|ccc}
        \hline
        Dataset & DIOR &DIOR-R &FAIR1M \\ 
        \hline
        \begin{tabular}[c]{@{}c@{}}Activated \\ modality \end{tabular} & HR & HR & HR  \\
        Optimizer  & AdamW & AdamW & AdamW \\
        Input size & 800$\times$800  & 800  $\times$800 & 512$\times$512  \\         
        Input channel & RGB & RGB & RGB \\ 
        Base lr.  & 8e-5 & 8e-5 &8e-5 \\
        Lr. scheduler & multistep & multistep & multistep   \\
        \begin{tabular}[c]{@{}c@{}}Layer-wise \\ lr decay\end{tabular}& 0.85   & 0.85    & 0.85 \\
        Weight decay & 0.05 & 0.05 & 0.05  \\
        Max epoch & 12 & 12  & 8 \\  
        Warmup  &linear &linear &linear    \\
        Warmup iters. & 1k & 1k  &0.5k  \\
        Warmup ratio & 1e-3 & 1e-3 & 1e-3 \\
        Drop path rate  &0.2 &0.2 &0.2 \\
        Augmentations & &  & \\
        RandomFlip &\checkmark &\checkmark &\checkmark \\
        RadnomRotate  &  &  &\checkmark \\
        Head &Faster RCNN &Oriented RCNN &Oriented RCNN \\
        \hline
        \end{tabular}
    \caption{The finetuning setting in object detection tasks. The probability of a random flip is 0.5.}
    \label{det_setting}
\end{table}

\subsection{Change Detection} 
Change detection focuses on identifying pixel-level regional changes using bi-temporal or multi-temporal images. Building upon the work of Sun et al. \cite{sun2022ringmo}, we incorporate the backbones of various RSFMs into the BIT framework \cite{chen2021remote} to evaluate their performance on the LEVIR-CD dataset. Following previous approaches \cite{SkySense,manas2021seasonal,mall2023change}, we utilize U-Net \cite{ronneberger2015u} as the segmentation head to assess the effectiveness of RSFMs in bi-temporal change detection tasks using the OSCD dataset with multi-spectral imagery. Additionally, we use the DynamicEarthNet-Sentinel2 dataset to evaluate model performance on semantic change detection tasks, maintaining the same configuration as the segmentation task. Further settings are detailed in Section \ref{cd_setting}.

\begin{table}
        \centering
        \setlength\tabcolsep{2pt}%
        \small
        \begin{tabular}{l|ccc}
        \hline
        Dataset & LEVIR-CD &OSCD &Dyna.-S2 \\ 
        \hline
        \begin{tabular}[c]{@{}c@{}}Activated \\ modality \end{tabular} & HR & MS & MS  \\
        Optimizer  & AdamW & AdamW & AdamW \\
        Input size & 256$\times$256  & 96  $\times$96 & 256$\times$256  \\         
        Input channel & RGB &  \begin{tabular}[c]{@{}c@{}}B02-08, B8A, \\ B11-12\end{tabular}  &  \begin{tabular}[c]{@{}c@{}}B02-08, B8A, \\ B11-12\end{tabular}  \\ 
        Base lr.  & 6e-5 & 6e-4 & 1e-4 \\
        Lr. scheduler & LambdaLR &ExponentialLR &poly   \\
        \begin{tabular}[c]{@{}c@{}}Layer-wise \\ lr decay\end{tabular}& 0.9   & 0.9    & 0.8 \\
        Weight decay & 0.01 & 1e-4 & 0.05  \\
        Max iters./epoch & 200 epochs &100 epochs &80k iters \\  
        Warmup  &- &- &linear    \\
        Warmup iters. & - & -  &1.5k  \\
        Warmup ratio & - & - & 1e-6 \\
        Drop path rate  &0.2 &0.2 &0.2 \\
        Augmentations & &  & \\
        RandomCrop &\checkmark & &\checkmark \\
        RandomFlip &\checkmark &\checkmark &\checkmark \\
        Head/Detector &BIT &U-Net &UperNet\\
        Loss &CrossEntropy &BCE &CrossEntropy \\

        \hline
        \end{tabular}
    \caption{The finetuning setting in change detection tasks.  The probability of a random flip is 0.5.}
    \label{cd_setting}
\end{table}

\subsection{Scene Classification} 
We select two widely-used single-label scene classification datasets: AID and NWPU-RESISC45. Additionally, we utilize a multi-label multispectral scene classification dataset, BigEarthNet-Sentinel2, and a temporal multispectral scene classification dataset, fMoW-Sentinel2. The AID and NWPU-RESISC45 (RESISC-45) datasets consist of high-resolution optical images, while BigEarthNet-Sentinel2 (BEN-S2) and fMoW-Sentinel2 (fMoW-S2) are extensive multispectral image datasets. Our scene classification experiments are carried out using a standard linear classifier. Detailed implementation settings can be found in Table~\ref{cls_setting}.

\begin{table}
        \centering
        \setlength\tabcolsep{2pt}%
        \small
        \begin{tabular}{l|cccc}
        \hline
        Dataset & AID &RESISC-45 &BEN-S2 &fMoW-S2\\ 
        \hline
        \begin{tabular}[c]{@{}c@{}}Activated \\ modality \end{tabular} & HR & HR & MS & MS \\
        Optimizer  & AdamW & AdamW & AdamW & AdamW    \\
        Input size & 320$\times$320  & 320$\times$320 & 128$\times$128   & 96$\times$96 \\         
        Input channel & RGB & RGB & \begin{tabular}[c]{@{}c@{}}B02-08, B8A, \\ B11-12\end{tabular}  & \begin{tabular}[c]{@{}c@{}}B02-08, B8A, \\ B11-12\end{tabular} \\ 
        Base lr.  & 6e-5 & 6e-5& 5e-5& 8e-4 \\
        Lr. scheduler & cosine & cosine & multistep & cosine  \\
        Weight decay & 0.05 &0.05 &0.01 &0.05  \\
        \begin{tabular}[c]{@{}c@{}}Layer-wise \\ lr decay\end{tabular}& 0.9   & 0.9   & 0.9   & 0.9  \\
        Max epoch &200 & 200 & 100 & 30    \\  
        Warmup  &linear &linear &- &linear    \\
        Warmup epoch & 5 & 5 & -  & 5  \\
        Warmup ratio & 0.01 & 0.01 & - & 0.2 \\
        Drop path rate  &0.2 &0.2 &0.2 &0.2 \\
        Augmentations & & & & \\
        RandomErasing &\checkmark &\checkmark & &\\
        RandomCrop &\checkmark &\checkmark & &\checkmark \\
        Mixup & & & &\checkmark \\
        RandomFlip &\checkmark &\checkmark &\checkmark &\checkmark \\
        \hline
        \end{tabular}
    \caption{The finetuning setting in single-modal semantical segmetation tasks. The minimum and maximum area ratio of random erasing are 0.03 and 0.333, respectively, and the probability of a random erasing is 0.3. The mixup ratio and probability are 0.8 and 1.0, respectively. The probability of a random flip is 0.5.}
    \label{cls_setting}
\end{table}

\begin{table*}
\centering
\setlength\tabcolsep{2.5pt}%
\small
\begin{tabular}{l|c|c|c}
\hline
Task                    & \begin{tabular}[c]{@{}c@{}}\texttt{(i)} Multi-Modal Segmentation: \\ Time-insensitive LandCover Mapping\end{tabular} & \begin{tabular}[c]{@{}c@{}}\texttt{(ii)} Multi-Modal Segmentation: \\ Time-sensitive Crop Mapping\end{tabular} & \multicolumn{1}{c}{\begin{tabular}[c]{@{}c@{}} \texttt{(iii)} Multi-Modal \\ Classification\end{tabular}} \\ \hline
Dataset                 & Dyna.-MM                                                                                      & PASTIS-MM                                                                                        & BEN-MM                                                                         \\ \hline
Optimizer               & AdamW                                                                                                         & AdamW                                                                                                 &   AdamW                                                                                         \\
Input Size              &   \makecell[c]{planet: 1024$\times$1024  \\ sentinel2: 1024$\times$1024 \\ sentinel1: 1024$\times$1024}                                                                                                       &   \makecell[c]{gep: 4096$\times$4096 \\ sentinel2: 128$\times$128 \\ sentinel1: 128$\times$128}                                                                                               &     \makecell[c]{sentinel2: 128$\times$128 \\ sentinel1: 128$\times$128}                                                                                        \\
Input channel           &        \makecell[c]{planet: RGBNIR \\ sentinel2: B02-08, B8A, B11-12 \\ sentinel1: VV, VH}                                                                                                 &    \makecell[c]{gep: RGB \\ sentinel2: B02-08, B8A, B11-12 \\ sentinel1: VV, VH}                                                                                              &     \makecell[c]{sentinel2: B02-08, B8A, B11-12 \\ sentinel1: VV, VH}                                                                                       \\
Base learning rate      &      6e-05                                                                                                   &     6e-05                                                                                             &     5e-05                                                                                      \\
Learning rate scheduler &     linear                                                                                                    &      linear                                                                                            &     MultiStepLR                                                                                       \\
Weight decay            &       0.01                                                                                                  &     0.01                                                                                             &     0.01                                                                                        \\
Batch size              &         8                                                                                                &      8                                                                                            &     256                                                                                       \\
Max iteration/epoch     &        6k iters                                                                                                 &      20k iters                                                                                            &     100 epoch                                                                                      \\
Warmup                  &      linear                                                                                                   &     linear                                                                                             &    -                                                                                       \\
Warmup iteration/epoch  &        150 iters                                                                                                 &    1500 iters                                                                                              &       -                                                                                   \\
Warmup ratio            &      1e-6                                                                                                   &    1e-6                                                                                              &      -                                                                                   \\
Drop path rate          &          0.2                                                                                               &      0.2                                                                                           &    0.2                                                                                       \\
Augmentation            &       \makecell[c]{RandomFlip}                                                                                                  &      \makecell[c]{RandomFlip}                                                                                            &            \makecell[c]{RandomFlip}                                                                                \\
Head/Detector            &       UperNet                                                                                                 &      FCN                                                                                            &     Linear Classifier                                                                                       \\
Loss function            &  CrossEntropy                                                                                                      &  CrossEntropy                                                                                                &   \makecell[c]{MultiLabel \\ SoftMargin}                                                                                  \\\hline
\end{tabular}
\caption{The finetuning setting in multi-modal downstream tasks.}
\label{multimodalimple}
\end{table*}

\subsection{Multi-Modal Semantic Segmentation} 
By integrating multi-modal data from a variety of sensors, imaging techniques, resolutions, and spectral bands, we can extract a richer and more distinctive set of features. These features improve the ability to understand and interpret the shape, size, and relationships among ground objects. To evaluate the tasks of Time-insensitive Land Cover Mapping and Time-sensitive Crop Mapping, we use the DynamicEarthNet-MM (Dyna.-MM) dataset and the PASTIS-MM dataset, respectively.
\paragraph{Dyna.-MM} contains spatially and temporally aligned multi-modal data, which include PlanetFusion imagery from the DynamicEarthNet-PlanetFusion dataset, Sentinel-2 multispectral imagery from the DynamicEarthNet-Sentinel2 dataset, and Sentinel-1 SAR imagery. For the SAR data, we utilize standard-calibrated Sentinel-1 GRD data with VV and VH polarizations, selecting it based on the geographical coordinates of the optical imagery. This approach is the same as SkySense \cite{SkySense} and ensures the validity of our multi-modal experiments. For segmentation tasks, UperNet is used as the segmentation head, and we report the mean Intersection over Union (mIoU) metric. Additional implementation details can be found in Table~\ref{multimodalimple} (\texttt{i}).
\paragraph{PASTIS-MM} \cite{SkySense,garnot2022multi} is a dataset sourced from SkySense\cite{SkySense}, which is designed for fine-grained, time-sensitive crop mapping. This dataset extends the PASTIS-R dataset \cite{garnot2022multi} by incorporating spatially aligned high-resolution RGB images. PASTIS-MM aims to explore the combined impact of high-resolution optical imagery, medium-resolution temporal multispectral data, and temporal synthetic aperture radar (SAR) data in the context of time-sensitive crop mapping. The dataset was collected based on geo-coordinates and acquisition dates from the image tiles of the original PASTIS-R dataset, sourced from \cite{SkySense}. PASTIS-MM comprises 2433 Sentinel-2 image tiles, each with dimensions of 128×128 pixels, 10 spectral bands, and a GSD of 10 meters. For each tile, the dataset includes all available Sentinel-2 and Sentinel-1 acquisition data from September 2018 to November 2019, along with additional high-resolution visible imagery. For segmentation, we employ a naive Fully Convolutional Network (FCN) head \cite{long2015fully} and report Overall Accuracy (OA) based on the official five-fold cross-validation of the dataset. Further implementation details can be found in Table~\ref{multimodalimple} (\texttt{ii}).

\subsection{Multi-Modal Scene Classification} 
Following SkySense \cite{SkySense}, we utilize the representative BigEarthNet-MM (BEN-MM) dataset to evaluate the performance of SkySense V2 in large-scale scene classification tasks, with a focus on integrating optical and SAR data. This dataset builds upon the BigEarthNet-Sentinel2 dataset by adding corresponding Sentinel-1 SAR data, thereby enabling the assessment of multi-label scene classification using both MS and SAR modalities. BEN-MM enriches each Sentinel-2 image patch from the BigEarthNet-Sentinel2 dataset with a preprocessed Sentinel-1 image patch taken around the same time. Each Sentinel-1 patch retains the annotation information from its corresponding Sentinel-2 patch and features a GSD of 10 meters. These patches provide dual-polarization information channels (VV and VH) and are collected in interferometric wide-swath mode. Consistent with prior studies \cite{SkySense, wang2023decur,fuller2023croma,wang2022ssl4eo}, we keep the same data splits as employed in the BigEarthNet-Sentinel2 dataset. Further implementation details can be found in Table~\ref{multimodalimple} (\texttt{iii}).

\section{Comparison of Parameter Numbers with SkySense}

\begin{table}[H]
        \centering
        \setlength\tabcolsep{2pt}%
        \small
        \begin{tabular}{l|c|c|c}
        \hline
        Model Name &SkySense &\begin{tabular}[c]{@{}c@{}}  SkySense V2 \\ w/o MoE \end{tabular} & SkySense V2\\ 
        \hline
        Tokenizer & \begin{tabular}[c]{@{}c@{}}  0.21M \\ \textcolor{cyan}{HR: 0.02} \\  \textcolor{cyan}{MS: 0.16}\\  \textcolor{cyan}{SAR: 0.03}\end{tabular} &  \begin{tabular}[c]{@{}c@{}}  0.09M \\ \textcolor{cyan}{HR: 0.02} \\  \textcolor{cyan}{MS: 0.06}\\  \textcolor{cyan}{SAR: 0.01}\end{tabular} &  \begin{tabular}[c]{@{}c@{}}  0.09M \\ \textcolor{cyan}{HR: 0.02} \\  \textcolor{cyan}{MS: 0.06}\\  \textcolor{cyan}{SAR: 0.01}\end{tabular}\\
        \hline
        Backbone &\begin{tabular}[c]{@{}c@{}}  1260.31M \\ \textcolor{cyan}{HR: 655.17} \\  \textcolor{cyan}{MS: 302.57}\\  \textcolor{cyan}{SAR: 302.57}\end{tabular} &661.40M &1994.10M \\
         \hline
        Modality prompt & - &9.94M &9.94M\\
        \hline
        Fusion module  &398.20M& 347.01M &347.01M  \\
        \hline
        Others  &404.13M& 490.49M &490.49M  \\
        \hline
        Total &2062.85M &1508.93M &2841.63M \\
        \hline
        \end{tabular}
    \caption{Comparison of the number of parameters in different modules between SkySense V2 and SkySense.}
    \label{param_cmp}
\end{table}

SkySense \cite{SkySense} employed three distinct backbones: Swin-H for high-resolution (HR) optical data, ViT-L for multi-spectral (MS) data, and ViT-L for synthetic aperture radar (SAR) data. In SkySense V2, the backbone parameters are shared across different modalities, maintaining a few separate parameters for modality-specific tokenizers and prompts. Detailed comparisons are presented in Table~\ref{param_cmp}. By adopting this unified design, the total number of backbone parameters for the three modalities has been reduced from 1,260 million to 661 million. Additionally, we incorporated a mixture of experts (MoE) approach \cite{MOE}, which allowed us to scale up the number of parameters to 1,994 million (with 661 million activated). To sum up, our unified transformer backbone employs full parameter sharing across different modalities, presenting several key benefits: 1) As discussed in the ablation part in our paper, this parameter sharing aggregates gradients from all modalities, thereby accelerating the convergence process. 2) It significantly boosts parameter utilization efficiency, leaving enough room for increasing additional capacity by incorporating MoE modules, which further enhances representation learning. 3) Our unified model architecture and complete parameter sharing simplifies the alignment of features across different modalities.

\section{Experiments}
\subsection{Influence of Image-text Alignment with OSM}
OpenStreetMap is a global open-source data providing a wealth of semantic classes. We utilize the CLIP text encoder \cite{radford2021learning} to transform categories into text representations and then apply dense image-text alignment (ITA) to enhance pre-trained model's capability for dense interpretation. To validate this approach, we conducted ablation experiments on segmentation datasets, specifically iSAID and Potsdam. Due to the resource-intensive nature of the whole pre-training, we ensured a fair comparison by limiting it to 20,000 iterations. The fine-tuning process was kept consistent with the approach outlined in Section \ref{semantic_segmentation}. The results, presented in Table~\ref{ablation_ita}, demonstrate that image-text alignment effectively improves the performance of dense tasks.

\begin{table}[h]
        \centering
        \setlength\tabcolsep{20pt}%
        \small
        \begin{tabular}{l|ccc}
        \hline
        Dataset &iSAID &Potsdam \\ 
        \hline

        w/o ITA &67.45 &88.77\\
        w/ ITA &68.24 &90.05\\

        \hline
        \end{tabular}
    \caption{Ablation results of image-text alignment in SkySense V2.}
    \label{ablation_ita}
\end{table}

\subsection{Features of Different Resolutions Derived from  Adaptive Patch Merging}
Our Adaptive Patch Merging (APM) module, integrated after each stage of the unified backbone, can flexibly generate features with various resolutions based on specific requirements. To evaluate the impact of different subsampling activation conditions within APM, we conducted ablation experiments on the segmentation datasets iSAID and Potsdam. The fine-tuning process remained consistent with the methodology outlined in Section \ref{semantic_segmentation}, and all models utilized parameters from the same pre-trained model. As shown in Table~\ref{ablation_apm}, generating higher-resolution features through APM enhances the model's performance. This improvement makes the model particularly advantageous for deployment in environments where sufficient computing resources are available.

    

\begin{table}[h]
        \centering
        \setlength\tabcolsep{2pt}%
        \small
        \begin{tabular}{ccc|c|cc}
        \hline
        \multicolumn{3}{c|}{Sub-sampling activation of APM} & downscale & \multicolumn{2}{c}{Dataset} \\
        Stage 2 \quad & Stage 3 & Stage 4 & &iSAID &Potsdam \\ 
        \hline
        \checkmark &\checkmark &\checkmark &$1 / 8$  &71.87 &95.86\\
        \checkmark &\checkmark & &$1 / 4$ &71.92 &95.85\\
        \checkmark & & &$1 / 2$ &72.55 &96.76 \\
         & & &$ - $ &72.88 &97.03\\
    
        \hline
        \end{tabular}
    \caption{Experiment results of different sub-sampling activation conditions within APM. All models were initialized with identical parameters, differing only in their subsampling activation strategies in APM. }
    \label{ablation_apm}
\end{table}

\subsection{Performance on Sensor Data Outside of Training}

To further validate the generalizability of the pre-trained model, we conducted experiments on three datasets collected from different sensors: Five-Billion-Pixels (FBP) \cite{FBP} from the Gaofen-2 satellite, SPARCS \cite{SPARCS} from the Landsat-8 satellite, and AIR-PolSAR-Seg (APS) \cite{Polsar} from the Gaofen-3 satellite. All these datasets utilize sensors different from those used in the training data. FBP comprises over 5 billion labeled pixels across 150 high-resolution images, annotated into 24 categories covering artificially constructed, agricultural, and natural classes. SPARCS includes 80 images with a resolution of $1000 \times 1000$ pixels, annotated into 7 categories. APS consists of a PolSAR image with a region of $9082 \times 9805$ pixels and 2000 image patches, each sized $512 \times 512$ pixels. The experimental results on these three datasets are presented in Table~\ref{ablation_general}. SkySense V2 surpasses SkySense by an average of 1.8\% in mIoU, indicating that SkySense V2 possesses stronger generalization capabilities than SkySense. We attribute this improvement to the unified design, which allows the backbone parameters to be trained with data from different modalities, thereby enhancing the model's ability to generalize effectively.

\begin{table}[h]
        \centering
        \setlength\tabcolsep{4pt}%
        \small
        \begin{tabular}{lc|cc}
        \hline
        Dataset &Sensor &SkySense & SkySense V2\\ 
        \hline
        Five-Billion-Pixels &Gaofen-2 &65.31 & 66.82 \\ 
        SPARCS &Landsat-8 & 72.57 &74.32 \\ 
        AIR-PolSAR-Seg &Gaofen-3(SAR) &53.21 & 55.32 \\ 
        \hline
        \end{tabular}
    \caption{Results on datasets built from various sensors. The evaluation metric is mIoU.}
    \label{ablation_general}
\end{table}

\subsection{Ablation of Modality-specific Prompt Tokens in Downstream Tasks}
After pre-training the model, there are two options for handling Modality-specific Prompt Tokens (MSPT) during downstream fine-tuning: 1) retaining the MSPT or 2) removing the MSPT entirely. We assess the impact of MSPT in two different settings: 1) single-modal tasks, where only one modality is activated, and 2) multi-modal tasks, where at least two modalities are activated. For the single-modal setting, we conduct experiments using the RESISC-45 and BEN-S2 datasets. For the multi-modal setting, we utilize the BEN-MM dataset. As demonstrated in Table~\ref{ablation_mspt}, our findings indicate that MSPT can significantly enhance performance in multi-modal tasks, primarily due to its ability to increase the diversity of features of different modalities.

\begin{table}[h]
        \centering
        \setlength\tabcolsep{5pt}%
        \small
        \begin{tabular}{l|cc|c}
        \hline
        Dataset &\begin{tabular}[c]{@{}c@{}} RESISC-45 \\ (TR=10\%) \end{tabular} & \begin{tabular}[c]{@{}c@{}} BEN-S2 \\ (TR=10\%) \end{tabular}  & BEN-MM  \\ 
        Activated modality &HR &MS &MS,SAR \\
        \hline
        w/o MSPT &96.15 &88.97& 92.64\\
        w/ MSPT &96.42 &89.13& 93.81\\
        \hline
        \end{tabular}
    \caption{Results of ablation study of modality-specific prompt token in downstream tasks. "TR" refers to training ratio, representing the proportion of training data relative to the entire dataset.}
    \label{ablation_mspt}
\end{table}

\subsection{Ablation Studies about MoE in Pre-training}
To quickly assess the impact of MOE-related configurations, we pre-trained the model with 20,000 iterations for each configuration. After pre-training, we evaluated the model on the AID and RESISC-45 datasets using the k-NN accuracy.
\paragraph{Varying the number of experts.} We configured the unified backbone with varying numbers of experts to evaluate performance relative to parameter size. The results, shown in Table~\ref{ablation_expert_num}, indicate that as the number of experts increases, the representational capacity of the model improves. Although the configuration with 16 experts outperforms that with 8 experts, it requires an additional 1.6 billion parameters. This increase in parameters does not match the marginal gain in performance. Consequently, we set the number of experts to 8 in our SkySense V2 model.
\begin{table}[h]
        \centering
        \setlength\tabcolsep{10pt}%
        \small
        \begin{tabular}{cc|ccc}
        \hline
         \#experts & \#parameters &AID &RESISC-45  \\ 
        \hline

        4& 1232.61M &89.05 &82.57\\
        8& 1994.10M &91.00 &85.11\\
        16& 3517.08M &91.23 &85.97 \\

        \hline
        \end{tabular}
    \caption{Ablation results of varing number of experts in MoE. We report k-NN classification accuracy on AID and RESISC-45 datasets.}
    \label{ablation_expert_num}
\end{table}

\paragraph{Varying the number of MoE blocks.} Following prior methods utilizing Mixture of Experts (MoE) \cite{ResidualMoE, Task-customizedMOE}, we integrate MoE modules into the last  $L$ transformer blocks, substituting the original feed-forward network (FFN) layers. Each MoE module comprises 8 experts, all of which maintain the FFN's structural design but function as independent networks. We present ablation studies exploring various configurations with different numbers of MoE blocks ($L=2,4,6,8$) within the backbone.  As shown in Table~\ref{ablation_moe_block}, the results indicate that performance tends to plateau at 6 MoE blocks.
\begin{table}[h]
        \centering
        \setlength\tabcolsep{10pt}%
        \small
        \begin{tabular}{cc|ccc}
        \hline
         $L$ & \#parameters &AID &RESISC-45  \\ 
        \hline

        2& 1486.37M &89.63 &83.45 \\
        4& 1740.23M &90.14 &84.37 \\
        6& 1994.10M &91.00 &85.11 \\
        8& 2247.97M &91.11 &85.43 \\

        \hline
        \end{tabular}
    \caption{Ablation results of varing number of MoE blocks in backbone.}
    \label{ablation_moe_block}
\end{table}

\paragraph{Exploring different distributions of MoE blocks in backbone.} Previous studies \cite{ResidualMoE, Task-customizedMOE} typically incorporate Mixture of Experts (MoE) into the last few layers of a network. This approach is motivated by two main factors: 1) deeper routing decisions are more closely related to image classes and contain richer semantic information \cite{Riquelme2021ScalingVW}, and 2) the last layers have the most significant impact on classification performance. However, the official implementation of Swin-MoE \cite{hwang2022tutel}\footnote{\url{https://github.com/microsoft/Swin-Transformer}} introduces an alternative strategy, distributing MoE blocks evenly across all layers in whole backbone. We tested both distribution strategies within the backbone of SkySense V2. Our findings indicate that while the performance difference between the two methods is minimal, incorporating MoE blocks into the last layers offers a slight advantage, as detailed in Table~\ref{ablation_moe_dist}.

\begin{table}[h]
        \centering
        \setlength\tabcolsep{10pt}%
        \small
        \begin{tabular}{c|cc}
        \hline
         \begin{tabular}[c]{@{}c@{}}  MoE block(layer) index \\ Total: 24 \end{tabular} &AID &RESISC-45  \\ 
        \hline
        \begin{tabular}[c]{@{}c@{}}  3,7,11, \\ 15,19,23 \end{tabular} &90.93 &84.87 \\
        \hline
        \begin{tabular}[c]{@{}c@{}}  18,19,20, \\ 21,22,23 \end{tabular} &91.00 &85.11 \\
        \hline
        \end{tabular}
    \caption{Comparison of different MoE distribution strategies within the backbone of SkySense V2.}
    \label{ablation_moe_dist}
\end{table}

\subsection{Ablation Studies about MoE in Downstream Tasks}

To further investigate the MoE, we conducted ablation experiments during the downstream fine-tuning phase of a pre-trained MoE backbone. We selected the RESISC-45, BEN-S2, and BEN-MM datasets, which encompass three modalities: high-resolution (HR), multispectral (MS), and synthetic aperture radar (SAR). Firstly, we examined whether to fix the routing gate during downstream fine-tuning. As shown in Table~\ref{ablation_moe_down}, despite the routing gate being trained with a substantial amount of data during the pre-training stage, fine-tuning for a specific task proves to be necessary. Secondly, we experimented by randomly keeping one expert from the MoE block while removing the others, effectively reducing the MoE feed-forward network (FFN) to a plain FFN. The performance of this modified model is comparable to the fully pre-trained backbone without MoE, indicating that each expert has been sufficiently trained and possesses individual representational capabilities.

\begin{table}[h]
        \centering
        \setlength\tabcolsep{4pt}%
        \small
        \begin{tabular}{c|cc|c}
        \hline
        Dataset &\begin{tabular}[c]{@{}c@{}} RESISC-45 \\ (TR=10\%) \end{tabular} & \begin{tabular}[c]{@{}c@{}} BEN-S2 \\ (TR=10\%) \end{tabular}  & BEN-MM  \\ 
        Activated modality &HR &MS &MS,SAR \\
        \hline
        SkySense V2 &96.42 &89.13& 93.81\\
         \hline
        SkySense V2 w/o MoE &95.61 &88.76& 92.95\\
         \hline
        \begin{tabular}[c]{@{}c@{}}  SkySense V2 \\ fixed routing gate \end{tabular} &95.73 &88.65& 92.80\\
         \hline

        \begin{tabular}[c]{@{}c@{}}  SkySense V2 \\ random 1 expert \end{tabular} &95.47 &88.69& 92.81\\
        \hline
        \end{tabular}
    \caption{Results of ablation study of MoE in downstream tasks. "TR" refers to training ratio, representing the proportion of training data relative to the entire dataset.}
    \label{ablation_moe_down}
\end{table}

\subsection{Ablation Studies about the Number of Queries in Query-based Semantic Aggregation Contrastive Learning}

In Query-based Semantic Aggregation Contrastive Learning (QSACL) ablation study, we explore the influence of different $m$ learnable queries, which are used to aggregate features with different semantics across multiple augmented views of an image. We pre-trained the model with 20,000 iterations, experimenting with various numbers of queries. After pre-training, we evaluated the model's performance on the AID and RESISC-45 datasets using k-NN accuracy. The results, as shown in Table~\ref{ablation_qsacl}, indicate that with a small number of queries, such as 4 or 8, performance drops significantly. This decline occurs because the number of queries is insufficient to capture the diverse semantic categories within an image, resulting in inadequate pre-training. Conversely, when $m=8$, the performance remains similar to when $m=16$, suggesting that 16 queries are sufficient to capture the different semantics in an image, with additional queries offering no significant improvement in performance.

\begin{table}[h]
        \centering
        \setlength\tabcolsep{20pt}%
        \small
        \begin{tabular}{c|ccc}
        \hline
         $m$ & AID &RESISC-45  \\ 
        \hline

        4& 90.21 &84.32 \\
        8& 90.68 &84.87 \\
        16& 91.00 &85.11 \\
        24& 91.05 &85.07 \\

        \hline
        \end{tabular}
    \caption{Ablation results of varing number of MoE learnable queries in QSACL.}
    \label{ablation_qsacl}
\end{table}

\subsection{Comparison with Random Initialization.}

In this section, we use both SkySense pre-trained weights and randomly initialized weights to fine-tune the same backbone network across three datasets, each corresponding to a different task. These tasks include scene classification with the AID dataset \cite{xia2017aid}, object detection with the DIOR dataset \cite{li2020object}, and semantic segmentation with the iSAID dataset \cite{waqas2019isaid}. The experimental results, which are shown in Table~\ref{randinit_vs_pretrain}, indicate a significant performance advantage for our pre-trained model compared to the model trained from scratch across all three datasets.

\begin{table}[h]
\centering
\begin{tabular}{cccc}
\toprule
\multirow{2}{*}{Model} & AID                 & DIOR    & iSAID \\ \cmidrule(lr){2-4} 
                       & OA(TR=20/50\%) & $\mathrm{mAP}_{50}$ & mIoU  \\ 
\midrule
Randm Init             & 66.82/90.78         & 56.36   & 48.34 \\
SkySense V2            & 98.34/99.05         & 79.50   & 71.87 \\ 
\bottomrule
\end{tabular}
\caption{Comparison of SkySense V2 with random initialization and SkySense V2 with pre-training.}
\label{randinit_vs_pretrain}
\end{table}

{
    \small
    \bibliographystyle{ieeenat_fullname}
    \bibliography{main}

\begin{thebibliography}{86}
\providecommand{\natexlab}[1]{#1}
\providecommand{\url}[1]{\texttt{#1}}
\expandafter\ifx\csname urlstyle\endcsname\relax
  \providecommand{\doi}[1]{doi: #1}\else
  \providecommand{\doi}{doi: \begingroup \urlstyle{rm}\Url}\fi

\bibitem[Akiva et~al.(2022)Akiva, Purri, and Leotta]{akiva2022self}
Peri Akiva, Matthew Purri, and Matthew Leotta.
\newblock Self-supervised material and texture representation learning for remote sensing tasks.
\newblock In \emph{Proceedings of the IEEE/CVF Conference on Computer Vision and Pattern Recognition}, pages 8203--8215, 2022.

\bibitem[Astruc et~al.(2024)Astruc, Gonthier, Mallet, and Landrieu]{AnySat}
Guillaume Astruc, Nicolas Gonthier, Clement Mallet, and Loic Landrieu.
\newblock {AnySat: An Earth} observation model for any resolutions, scales, and modalities.
\newblock \emph{arXiv preprint arXiv:2412.14123}, 2024.

\bibitem[Ayush et~al.(2021)Ayush, Uzkent, Meng, Tanmay, Burke, Lobell, and Ermon]{ayush2021geography}
Kumar Ayush, Burak Uzkent, Chenlin Meng, Kumar Tanmay, Marshall Burke, David Lobell, and Stefano Ermon.
\newblock Geography-aware self-supervised learning.
\newblock In \emph{Proceedings of the IEEE/CVF International Conference on Computer Vision}, pages 10181--10190, 2021.

\bibitem[Bastani et~al.(2023)Bastani, Wolters, Gupta, Ferdinando, and Kembhavi]{bastani2022satlas}
Favyen Bastani, Piper Wolters, Ritwik Gupta, Joe Ferdinando, and Aniruddha Kembhavi.
\newblock Satlaspretrain: A large-scale dataset for remote sensing image understanding.
\newblock \emph{Proceedings of the IEEE/CVF International Conference on Computer Vision}, pages 16772--16782, 2023.

\bibitem[Cao et~al.(2023)Cao, Huang, and Weng]{cao2023multi}
Yinxia Cao, Xin Huang, and Qihao Weng.
\newblock A multi-scale weakly supervised learning method with adaptive online noise correction for high-resolution change detection of built-up areas.
\newblock \emph{Remote Sensing of Environment}, 297:\penalty0 113779, 2023.

\bibitem[Caron et~al.(2020)Caron, Misra, Mairal, Goyal, Bojanowski, and Joulin]{caron2020unsupervised}
Mathilde Caron, Ishan Misra, Julien Mairal, Priya Goyal, Piotr Bojanowski, and Armand Joulin.
\newblock Unsupervised learning of visual features by contrasting cluster assignments.
\newblock \emph{Advances in neural information processing systems}, 33:\penalty0 9912--9924, 2020.

\bibitem[Caron et~al.(2021)Caron, Touvron, Misra, J{\'e}gou, Mairal, Bojanowski, and Joulin]{caron2021emerging}
Mathilde Caron, Hugo Touvron, Ishan Misra, Herv{\'e} J{\'e}gou, Julien Mairal, Piotr Bojanowski, and Armand Joulin.
\newblock Emerging properties in self-supervised vision transformers.
\newblock In \emph{Proceedings of the IEEE/CVF international conference on computer vision}, pages 9650--9660, 2021.

\bibitem[Cha et~al.(2023)Cha, Seo, and Lee]{cha2023billion}
Keumgang Cha, Junghoon Seo, and Taekyung Lee.
\newblock A billion-scale foundation model for remote sensing images.
\newblock \emph{arXiv preprint arXiv:2304.05215}, 2023.

\bibitem[Chen and Shi(2020)]{chen2020spatial}
Hao Chen and Zhenwei Shi.
\newblock A spatial-temporal attention-based method and a new dataset for remote sensing image change detection.
\newblock \emph{Remote Sensing}, 12\penalty0 (10):\penalty0 1662, 2020.

\bibitem[Chen et~al.(2021)Chen, Qi, and Shi]{chen2021remote}
Hao Chen, Zipeng Qi, and Zhenwei Shi.
\newblock Remote sensing image change detection with transformers.
\newblock \emph{IEEE Transactions on Geoscience and Remote Sensing}, 60:\penalty0 1--14, 2021.

\bibitem[Chen et~al.(2019)Chen, Jiang, Wang, Cui, and Qian]{chen2019collaborative}
Wuyang Chen, Ziyu Jiang, Zhangyang Wang, Kexin Cui, and Xiaoning Qian.
\newblock Collaborative global-local networks for memory-efficient segmentation of ultra-high resolution images.
\newblock In \emph{Proceedings of the IEEE/CVF conference on computer vision and pattern recognition}, pages 8924--8933, 2019.

\bibitem[Cheng and Han(2016)]{cheng2016survey}
Gong Cheng and Junwei Han.
\newblock A survey on object detection in optical remote sensing images.
\newblock \emph{ISPRS journal of photogrammetry and remote sensing}, 117:\penalty0 11--28, 2016.

\bibitem[Cheng et~al.(2017)Cheng, Han, and Lu]{cheng2017remote}
Gong Cheng, Junwei Han, and Xiaoqiang Lu.
\newblock Remote sensing image scene classification: Benchmark and state of the art.
\newblock \emph{Proceedings of the IEEE}, 105\penalty0 (10):\penalty0 1865--1883, 2017.

\bibitem[Cheng et~al.(2022)Cheng, Wang, Li, Xie, Lang, Yao, and Han]{cheng2022anchor}
Gong Cheng, Jiabao Wang, Ke Li, Xingxing Xie, Chunbo Lang, Yanqing Yao, and Junwei Han.
\newblock Anchor-free oriented proposal generator for object detection.
\newblock \emph{IEEE Transactions on Geoscience and Remote Sensing}, 60:\penalty0 1--11, 2022.

\bibitem[Cong et~al.(2022)Cong, Khanna, Meng, Liu, Rozi, He, Burke, Lobell, and Ermon]{cong2022satmae}
Yezhen Cong, Samar Khanna, Chenlin Meng, Patrick Liu, Erik Rozi, Yutong He, Marshall Burke, David Lobell, and Stefano Ermon.
\newblock Satmae: Pre-training transformers for temporal and multi-spectral satellite imagery.
\newblock \emph{Advances in Neural Information Processing Systems}, 35:\penalty0 197--211, 2022.

\bibitem[Darcet et~al.(2024)Darcet, Oquab, Mairal, and Bojanowski]{ViTR}
Timoth{\'e}e Darcet, Maxime Oquab, Julien Mairal, and Piotr Bojanowski.
\newblock Vision transformers need registers.
\newblock In \emph{The Twelfth International Conference on Learning Representations}, 2024.

\bibitem[Daudt et~al.(2018)Daudt, Le~Saux, Boulch, and Gousseau]{daudt2018urban}
Rodrigo~Caye Daudt, Bertr Le~Saux, Alexandre Boulch, and Yann Gousseau.
\newblock Urban change detection for multispectral earth observation using convolutional neural networks.
\newblock In \emph{IGARSS 2018-2018 IEEE International Geoscience and Remote Sensing Symposium}, pages 2115--2118. Ieee, 2018.

\bibitem[Dosovitskiy et~al.(2020)Dosovitskiy, Beyer, Kolesnikov, Weissenborn, Zhai, Unterthiner, Dehghani, Minderer, Heigold, Gelly, et~al.]{dosovitskiy2020image}
Alexey Dosovitskiy, Lucas Beyer, Alexander Kolesnikov, Dirk Weissenborn, Xiaohua Zhai, Thomas Unterthiner, Mostafa Dehghani, Matthias Minderer, Georg Heigold, Sylvain Gelly, et~al.
\newblock An image is worth 16x16 words: Transformers for image recognition at scale.
\newblock \emph{arXiv preprint arXiv:2010.11929}, 2020.

\bibitem[Fedus et~al.(2021)Fedus, Zoph, and Shazeer]{SwitchTS}
William Fedus, Barret Zoph, and Noam~M. Shazeer.
\newblock Switch transformers: Scaling to trillion parameter models with simple and efficient sparsity.
\newblock \emph{ArXiv}, abs/2101.03961, 2021.

\bibitem[Fuller et~al.(2023)Fuller, Millard, and Green]{fuller2023croma}
Anthony Fuller, Koreen Millard, and James~R. Green.
\newblock Croma: Remote sensing representations with contrastive radar-optical masked autoencoders.
\newblock \emph{Advances in Neural Information Processing Systems}, 2023.

\bibitem[Garnot et~al.(2022)Garnot, Landrieu, and Chehata]{garnot2022multi}
Vivien Sainte~Fare Garnot, Loic Landrieu, and Nesrine Chehata.
\newblock Multi-modal temporal attention models for crop mapping from satellite time series.
\newblock \emph{ISPRS Journal of Photogrammetry and Remote Sensing}, 187:\penalty0 294--305, 2022.

\bibitem[Grill et~al.(2020)Grill, Strub, Altch{\'e}, Tallec, Richemond, Buchatskaya, Doersch, Avila~Pires, Guo, Gheshlaghi~Azar, et~al.]{grill2020bootstrap}
Jean-Bastien Grill, Florian Strub, Florent Altch{\'e}, Corentin Tallec, Pierre Richemond, Elena Buchatskaya, Carl Doersch, Bernardo Avila~Pires, Zhaohan Guo, Mohammad Gheshlaghi~Azar, et~al.
\newblock Bootstrap your own latent-a new approach to self-supervised learning.
\newblock \emph{Advances in neural information processing systems}, 33:\penalty0 21271--21284, 2020.

\bibitem[Guo et~al.(2022)Guo, Liu, Gan, Wang, Zhang, Wang, Jiang, Zhang, Yi, Ma, et~al.]{guo2022isdnet}
Shaohua Guo, Liang Liu, Zhenye Gan, Yabiao Wang, Wuhao Zhang, Chengjie Wang, Guannan Jiang, Wei Zhang, Ran Yi, Lizhuang Ma, et~al.
\newblock Isdnet: Integrating shallow and deep networks for efficient ultra-high resolution segmentation.
\newblock In \emph{Proceedings of the IEEE/CVF Conference on Computer Vision and Pattern Recognition}, pages 4361--4370, 2022.

\bibitem[Guo et~al.(2024)Guo, Lao, Dang, Zhang, Yu, Ru, Zhong, Huang, Wu, Hu, He, Wang, Chen, Yang, Zhang, and Li]{SkySense}
Xin Guo, Jiangwei Lao, Bo Dang, Yingying Zhang, Lei Yu, Lixiang Ru, Liheng Zhong, Ziyuan Huang, Kang Wu, Dingxiang Hu, Huimei He, Jian Wang, Jingdong Chen, Ming Yang, Yongjun Zhang, and Yansheng Li.
\newblock Skysense: A multi-modal remote sensing foundation model towards universal interpretation for earth observation imagery.
\newblock In \emph{Proceedings of the IEEE/CVF Conference on Computer Vision and Pattern Recognition (CVPR)}, pages 27672--27683, 2024.

\bibitem[Haiyang et~al.(2023)Haiyang, Hao, Shaoshuai, Aoxue, Zhenguo, Bernt, and Wang]{UniTR}
Wang Haiyang, Tang Hao, Shi Shaoshuai, Li Aoxue, Li Zhenguo, Schiele Bernt, and Liwei Wang.
\newblock Unitr: A unified and efficient multi-modal transformer for bird's-eye-view representation.
\newblock In \emph{ICCV}, 2023.

\bibitem[Han et~al.(2024)Han, Zhang, Shi, and Reichstein]{msGFM}
Boran Han, Shuai Zhang, Xingjian Shi, and Markus Reichstein.
\newblock Bridging remote sensors with multisensor geospatial foundation models.
\newblock \emph{2024 IEEE/CVF Conference on Computer Vision and Pattern Recognition (CVPR)}, pages 27852--27862, 2024.

\bibitem[He et~al.(2020)He, Fan, Wu, Xie, and Girshick]{he2020momentum}
Kaiming He, Haoqi Fan, Yuxin Wu, Saining Xie, and Ross Girshick.
\newblock Momentum contrast for unsupervised visual representation learning.
\newblock In \emph{Proceedings of the IEEE/CVF conference on computer vision and pattern recognition}, pages 9729--9738, 2020.

\bibitem[Hong et~al.(2024)Hong, Zhang, Li, Li, Li, Yao, Ghamisi, Yokoya, Li, Jia, Plaza, Gamba, Benediktsson, and Chanussot]{hong2024spectralgpt}
Danfeng Hong, Bing Zhang, Xuyang Li, Yuxuan Li, Chenyu Li, Jing Yao, Pedram Ghamisi, Naoto Yokoya, Hao Li, Xiuping Jia, Antonio Plaza, Paolo Gamba, Jon~Atli Benediktsson, and Jocelyn Chanussot.
\newblock Spectralgpt: Spectral remote sensing foundation model.
\newblock \emph{IEEE Transactions on Pattern Analysis and Machine Intelligence}, \penalty0 (8):\penalty0 5227--5244, 2024.

\bibitem[Hu et~al.(2020)Hu, Mou, and Zhu]{hu2020unsupervised}
Jingliang Hu, Lichao Mou, and Xiao~Xiang Zhu.
\newblock Unsupervised domain adaptation using a teacher-student network for cross-city classification of sentinel-2 images.
\newblock \emph{The International Archives of the Photogrammetry, Remote Sensing and Spatial Information Sciences}, 43:\penalty0 1569--1574, 2020.

\bibitem[Huang et~al.(2022)Huang, Song, Yang, Wang, Ren, Dong, Feng, Yin, and Li]{huang2022toward}
Xin Huang, Yihong Song, Jie Yang, Wenrui Wang, Huiqun Ren, Mengjie Dong, Yujin Feng, Haidan Yin, and Jiayi Li.
\newblock Toward accurate mapping of 30-m time-series global impervious surface area (gisa).
\newblock \emph{International Journal of Applied Earth Observation and Geoinformation}, 109:\penalty0 102787, 2022.

\bibitem[Hughes and Kennedy(2019)]{SPARCS}
M.~Joseph Hughes and Robert~H. Kennedy.
\newblock High-quality cloud masking of landsat 8 imagery using convolutional neural networks.
\newblock \emph{Remote. Sens.}, 11:\penalty0 2591, 2019.

\bibitem[Hwang et~al.(2022)Hwang, Cui, Xiong, Yang, Liu, Hu, Wang, Salas, Jose, Ram, Chau, Cheng, Yang, Yang, and Xiong]{hwang2022tutel}
Changho Hwang, Wei Cui, Yifan Xiong, Ziyue Yang, Ze Liu, Han Hu, Zilong Wang, Rafael Salas, Jithin Jose, Prabhat Ram, Joe Chau, Peng Cheng, Fan Yang, Mao Yang, and Yongqiang Xiong.
\newblock Tutel: Adaptive mixture-of-experts at scale, 2022.

\bibitem[Jacobs et~al.(1991)Jacobs, Jordan, Nowlan, and Hinton]{MOE}
Robert~A. Jacobs, Michael~I. Jordan, Steven~J. Nowlan, and Geoffrey~E. Hinton.
\newblock Adaptive mixtures of local experts.
\newblock \emph{Neural Computation}, 3:\penalty0 79--87, 1991.

\bibitem[Jain et~al.(2022)Jain, Schoen-Phelan, and Ross]{RS_BYOL}
Pallavi Jain, Bianca Schoen-Phelan, and Robert~J. Ross.
\newblock Self-supervised learning for invariant representations from multi-spectral and sar images.
\newblock \emph{IEEE Journal of Selected Topics in Applied Earth Observations and Remote Sensing}, 15:\penalty0 7797--7808, 2022.

\bibitem[Jia et~al.(2022)Jia, Tang, Chen, Cardie, Belongie, Hariharan, and Lim]{VPT}
Menglin Jia, Luming Tang, Bor-Chun Chen, Claire Cardie, Serge Belongie, Bharath Hariharan, and Ser-Nam Lim.
\newblock Visual prompt tuning.
\newblock In \emph{European Conference on Computer Vision (ECCV)}, 2022.

\bibitem[Li et~al.(2020)Li, Wan, Cheng, Meng, and Han]{li2020object}
Ke Li, Gang Wan, Gong Cheng, Liqiu Meng, and Junwei Han.
\newblock Object detection in optical remote sensing images: A survey and a new benchmark.
\newblock \emph{ISPRS journal of photogrammetry and remote sensing}, 159:\penalty0 296--307, 2020.

\bibitem[Li et~al.(2022)Li, Chen, Hu, and Zhu]{li2022oriented}
Wentong Li, Yijie Chen, Kaixuan Hu, and Jianke Zhu.
\newblock Oriented reppoints for aerial object detection.
\newblock In \emph{Proceedings of the IEEE/CVF conference on computer vision and pattern recognition}, pages 1829--1838, 2022.

\bibitem[Li et~al.(2024{\natexlab{a}})Li, Hong, and Chanussot]{Li_2024_CVPR}
Xuyang Li, Danfeng Hong, and Jocelyn Chanussot.
\newblock S2mae: A spatial-spectral pretraining foundation model for spectral remote sensing data.
\newblock In \emph{Proceedings of the IEEE/CVF Conference on Computer Vision and Pattern Recognition (CVPR)}, pages 24088--24097, 2024{\natexlab{a}}.

\bibitem[Li et~al.(2024{\natexlab{b}})Li, Hou, Ma, Wu, Guo, Ren, and Jiao]{MA3E}
Zhihao Li, Biao Hou, Siteng Ma, Zitong Wu, Xianpeng Guo, Bo Ren, and Licheng Jiao.
\newblock Masked angle-aware autoencoder for remote sensing images.
\newblock In \emph{European Conference on Computer Vision}, pages 260--278. Springer, 2024{\natexlab{b}}.

\bibitem[Liu et~al.(2023)Liu, Shi, Wang, and Zhong]{liu2023seeing}
Yinhe Liu, Sunan Shi, Junjue Wang, and Yanfei Zhong.
\newblock Seeing beyond the patch: Scale-adaptive semantic segmentation of high-resolution remote sensing imagery based on reinforcement learning.
\newblock In \emph{Proceedings of the IEEE/CVF International Conference on Computer Vision}, pages 16868--16878, 2023.

\bibitem[Liu et~al.(2021)Liu, Lin, Cao, Hu, Wei, Zhang, Lin, and Guo]{liu2021swin}
Ze Liu, Yutong Lin, Yue Cao, Han Hu, Yixuan Wei, Zheng Zhang, Stephen Lin, and Baining Guo.
\newblock Swin transformer: Hierarchical vision transformer using shifted windows.
\newblock In \emph{Proceedings of the IEEE/CVF international conference on computer vision}, pages 10012--10022, 2021.

\bibitem[Liu et~al.(2022)Liu, Hu, Lin, Yao, Xie, Wei, Ning, Cao, Zhang, Dong, Wei, and Guo]{Swinv2}
Ze Liu, Han Hu, Yutong Lin, Zhuliang Yao, Zhenda Xie, Yixuan Wei, Jia Ning, Yue Cao, Zheng Zhang, Li Dong, Furu Wei, and Baining Guo.
\newblock Swin transformer v2: Scaling up capacity and resolution.
\newblock In \emph{Proceedings of the IEEE/CVF Conference on Computer Vision and Pattern Recognition (CVPR)}, pages 12009--12019, 2022.

\bibitem[Liu et~al.(2024)Liu, Chen, Han, Hong, Xu, Li, and Kwok]{Task-customizedMOE}
Zhili Liu, Kai Chen, Jianhua Han, Lanqing Hong, Hang Xu, Zhenguo Li, and James Tin-Yau Kwok.
\newblock Task-customized masked autoencoder via mixture of cluster-conditional experts.
\newblock \emph{ArXiv}, abs/2402.05382, 2024.

\bibitem[Long et~al.(2015)Long, Shelhamer, and Darrell]{long2015fully}
Jonathan Long, Evan Shelhamer, and Trevor Darrell.
\newblock Fully convolutional networks for semantic segmentation.
\newblock In \emph{Proceedings of the IEEE conference on computer vision and pattern recognition}, pages 3431--3440, 2015.

\bibitem[Loshchilov and Hutter(2016)]{SGDRSG}
Ilya Loshchilov and Frank Hutter.
\newblock Sgdr: Stochastic gradient descent with warm restarts.
\newblock \emph{arXiv: Learning}, 2016.

\bibitem[Loshchilov and Hutter(2017)]{loshchilov2018fixing}
Ilya Loshchilov and Frank Hutter.
\newblock Fixing weight decay regularization in adam.
\newblock \emph{CoRR}, abs/1711.05101, 2017.

\bibitem[Lv et~al.(2022)Lv, Huang, Li, Zhao, Benediktsson, Sun, and Falco]{lv2022land}
ZhiYong Lv, HaiTao Huang, Xinghua Li, MingHua Zhao, Jon~Atli Benediktsson, WeiWei Sun, and Nicola Falco.
\newblock Land cover change detection with heterogeneous remote sensing images: Review, progress, and perspective.
\newblock \emph{Proceedings of the IEEE}, 2022.

\bibitem[Mall et~al.(2023)Mall, Hariharan, and Bala]{mall2023change}
Utkarsh Mall, Bharath Hariharan, and Kavita Bala.
\newblock Change-aware sampling and contrastive learning for satellite images.
\newblock In \emph{Proceedings of the IEEE/CVF Conference on Computer Vision and Pattern Recognition}, pages 5261--5270, 2023.

\bibitem[Manas et~al.(2021)Manas, Lacoste, Gir{\'o}-i Nieto, Vazquez, and Rodriguez]{manas2021seasonal}
Oscar Manas, Alexandre Lacoste, Xavier Gir{\'o}-i Nieto, David Vazquez, and Pau Rodriguez.
\newblock Seasonal contrast: Unsupervised pre-training from uncurated remote sensing data.
\newblock In \emph{Proceedings of the IEEE/CVF International Conference on Computer Vision}, pages 9414--9423, 2021.

\bibitem[Mendieta et~al.(2023)Mendieta, Han, Shi, Zhu, Chen, and Li]{mendieta2023gfm}
Mat{\'\i}as Mendieta, Boran Han, Xingjian Shi, Yi Zhu, Chen Chen, and Mu Li.
\newblock Towards geospatial foundation models via continual pretraining.
\newblock \emph{Proceedings of the IEEE/CVF International Conference on Computer Vision}, pages 16806--16816, 2023.

\bibitem[Muhtar et~al.(2023)Muhtar, Zhang, Xiao, Li, and Gu]{muhtar2023cmid}
Dilxat Muhtar, Xueliang Zhang, Pengfeng Xiao, Zhenshi Li, and Feng Gu.
\newblock Cmid: A unified self-supervised learning framework for remote sensing image understanding.
\newblock \emph{IEEE Transactions on Geoscience and Remote Sensing}, 2023.

\bibitem[Noman et~al.(2024)Noman, Naseer, Cholakkal, Anwer, Khan, and Khan]{SatMAE++}
Mubashir Noman, Muzammal Naseer, Hisham Cholakkal, Rao~Muhammad Anwer, Salman Khan, and Fahad~Shahbaz Khan.
\newblock Rethinking transformers pre-training for multi-spectral satellite imagery.
\newblock In \emph{Proceedings of the IEEE/CVF Conference on Computer Vision and Pattern Recognition (CVPR)}, pages 27811--27819, 2024.

\bibitem[Oquab et~al.(2023)Oquab, Darcet, Moutakanni, Vo, Szafraniec, Khalidov, Fernandez, Haziza, Massa, El-Nouby, et~al.]{oquab2023dinov2}
Maxime Oquab, Timoth{\'e}e Darcet, Th{\'e}o Moutakanni, Huy Vo, Marc Szafraniec, Vasil Khalidov, Pierre Fernandez, Daniel Haziza, Francisco Massa, Alaaeldin El-Nouby, et~al.
\newblock Dinov2: Learning robust visual features without supervision.
\newblock \emph{arXiv preprint arXiv:2304.07193}, 2023.

\bibitem[Radford et~al.(2021)Radford, Kim, Hallacy, Ramesh, Goh, Agarwal, Sastry, Askell, Mishkin, Clark, et~al.]{radford2021learning}
Alec Radford, Jong~Wook Kim, Chris Hallacy, Aditya Ramesh, Gabriel Goh, Sandhini Agarwal, Girish Sastry, Amanda Askell, Pamela Mishkin, Jack Clark, et~al.
\newblock Learning transferable visual models from natural language supervision.
\newblock In \emph{International conference on machine learning}, pages 8748--8763. PMLR, 2021.

\bibitem[Reed et~al.(2023)Reed, Gupta, Li, Brockman, Funk, Clipp, Keutzer, Candido, Uyttendaele, and Darrell]{reed2022scale}
Colorado~J Reed, Ritwik Gupta, Shufan Li, Sarah Brockman, Christopher Funk, Brian Clipp, Kurt Keutzer, Salvatore Candido, Matt Uyttendaele, and Trevor Darrell.
\newblock Scale-mae: A scale-aware masked autoencoder for multiscale geospatial representation learning.
\newblock In \emph{Proceedings of the IEEE/CVF International Conference on Computer Vision}, pages 4088--4099, 2023.

\bibitem[Ren et~al.(2015)Ren, He, Girshick, and Sun]{ren2015faster}
Shaoqing Ren, Kaiming He, Ross Girshick, and Jian Sun.
\newblock Faster r-cnn: Towards real-time object detection with region proposal networks.
\newblock \emph{Advances in neural information processing systems}, 28, 2015.

\bibitem[Riquelme et~al.(2021)Riquelme, Puigcerver, Mustafa, Neumann, Jenatton, Pinto, Keysers, and Houlsby]{Riquelme2021ScalingVW}
Carlos Riquelme, Joan Puigcerver, Basil Mustafa, Maxim Neumann, Rodolphe Jenatton, Andr{\'e}~Susano Pinto, Daniel Keysers, and Neil Houlsby.
\newblock Scaling vision with sparse mixture of experts.
\newblock In \emph{Neural Information Processing Systems}, 2021.

\bibitem[Ronneberger et~al.(2015)Ronneberger, Fischer, and Brox]{ronneberger2015u}
Olaf Ronneberger, Philipp Fischer, and Thomas Brox.
\newblock U-net: Convolutional networks for biomedical image segmentation.
\newblock In \emph{Medical Image Computing and Computer-Assisted Intervention--MICCAI 2015: 18th International Conference, Munich, Germany, October 5-9, 2015, Proceedings, Part III 18}, pages 234--241. Springer, 2015.

\bibitem[Shen et~al.(2023)Shen, Hou, Zhou, Du, Longpre, Wei, Chung, Zoph, Fedus, Chen, Vu, Wu, Chen, Webson, Li, Zhao, Yu, Keutzer, Darrell, and Zhou]{MOE_IT}
Sheng Shen, Le Hou, Yan-Quan Zhou, Nan Du, S. Longpre, Jason Wei, Hyung~Won Chung, Barret Zoph, William Fedus, Xinyun Chen, Tu Vu, Yuexin Wu, Wuyang Chen, Albert Webson, Yunxuan Li, Vincent Zhao, Hongkun Yu, Kurt Keutzer, Trevor Darrell, and Denny Zhou.
\newblock Mixture-of-experts meets instruction tuning: A winning combination for large language models.
\newblock In \emph{International Conference on Learning Representations}, 2023.

\bibitem[Shen et~al.(2024)Shen, Hou, Zhou, Du, Longpre, Wei, Chung, Zoph, Fedus, Chen, Vu, Wu, Chen, Webson, Li, Zhao, Yu, Keutzer, Darrell, and Zhou]{VLMO}
Sheng Shen, Le Hou, Yanqi Zhou, Nan Du, Shayne Longpre, Jason Wei, Hyung~Won Chung, Barret Zoph, William Fedus, Xinyun Chen, Tu Vu, Yuexin Wu, Wuyang Chen, Albert Webson, Yunxuan Li, Vincent~Y Zhao, Hongkun Yu, Kurt Keutzer, Trevor Darrell, and Denny Zhou.
\newblock Mixture-of-experts meets instruction tuning: A winning combination for large language models.
\newblock In \emph{The Twelfth International Conference on Learning Representations}, 2024.

\bibitem[Sherrah(2016)]{sherrah2016fully}
Jamie Sherrah.
\newblock Fully convolutional networks for dense semantic labelling of high-resolution aerial imagery.
\newblock \emph{arXiv preprint arXiv:1606.02585}, 2016.

\bibitem[Sumbul et~al.(2019)Sumbul, Kang, Kreuziger, Marcelino, Costa, Benevides, Caetano, and Demir]{sumbul2020bigearthnet}
Gencer Sumbul, Jian Kang, Tristan Kreuziger, Filipe Marcelino, Hugo Costa, Pedro Benevides, Mario Caetano, and Beg{\"u}m Demir.
\newblock Bigearthnet: A large-scale benchmark archive for remote sensing image understanding.
\newblock In \emph{IEEE International Geoscience and Remote Sensing Symposium}, pages 5901--5904, 2019.

\bibitem[Sumbul et~al.(2021)Sumbul, Wall, Kreuziger, Marcelino, Costa, Benevides, Caetano, Demir, and Mark]{Sumbul2021bigearthnet}
Gencer Sumbul, Arne~de Wall, Tristan Kreuziger, Filipe Marcelino, Hugo Costa, Pedro Benevides, Mario Caetano, Beg{\"u}m Demir, and Volkerl Mark.
\newblock {BigEarthNet}-{MM}: A large-scale, multimodal, multilabel benchmark archive for remote sensing image classification and retrieval.
\newblock \emph{{IEEE} Geoscience and Remote Sensing Magazine}, 9\penalty0 (3):\penalty0 174--180, 2021.

\bibitem[Sun et~al.(2022{\natexlab{a}})Sun, Wang, Lu, Zhu, Lu, He, Li, Rong, Yang, Chang, et~al.]{sun2022ringmo}
Xian Sun, Peijin Wang, Wanxuan Lu, Zicong Zhu, Xiaonan Lu, Qibin He, Junxi Li, Xuee Rong, Zhujun Yang, Hao Chang, et~al.
\newblock Ringmo: A remote sensing foundation model with masked image modeling.
\newblock \emph{IEEE Transactions on Geoscience and Remote Sensing}, 2022{\natexlab{a}}.

\bibitem[Sun et~al.(2022{\natexlab{b}})Sun, Wang, Yan, Xu, Wang, Diao, Chen, Li, Feng, Xu, et~al.]{sun2022fair1m}
Xian Sun, Peijin Wang, Zhiyuan Yan, Feng Xu, Ruiping Wang, Wenhui Diao, Jin Chen, Jihao Li, Yingchao Feng, Tao Xu, et~al.
\newblock Fair1m: A benchmark dataset for fine-grained object recognition in high-resolution remote sensing imagery.
\newblock \emph{ISPRS Journal of Photogrammetry and Remote Sensing}, 184:\penalty0 116--130, 2022{\natexlab{b}}.

\bibitem[Tao et~al.(2023)Tao, Qi, Zhang, Zhu, Lu, and Li]{tao2023tov}
Chao Tao, Ji Qi, Guo Zhang, Qing Zhu, Weipeng Lu, and Haifeng Li.
\newblock Tov: The original vision model for optical remote sensing image understanding via self-supervised learning.
\newblock \emph{IEEE Journal of Selected Topics in Applied Earth Observations and Remote Sensing}, 2023.

\bibitem[Toker et~al.(2022)Toker, Kondmann, Weber, Eisenberger, Camero, Hu, Hoderlein, {\c{S}}enaras, Davis, Cremers, et~al.]{toker2022dynamicearthnet}
Aysim Toker, Lukas Kondmann, Mark Weber, Marvin Eisenberger, Andr{\'e}s Camero, Jingliang Hu, Ariadna~Pregel Hoderlein, {\c{C}}a{\u{g}}lar {\c{S}}enaras, Timothy Davis, Daniel Cremers, et~al.
\newblock Dynamicearthnet: Daily multi-spectral satellite dataset for semantic change segmentation.
\newblock In \emph{Proceedings of the IEEE/CVF Conference on Computer Vision and Pattern Recognition}, pages 21158--21167, 2022.

\bibitem[Tong et~al.(2022)Tong, Xia, and Zhu]{FBP}
Xin-Yi Tong, Guisong Xia, and Xiaoxiang Zhu.
\newblock Enabling country-scale land cover mapping with meter-resolution satellite imagery.
\newblock \emph{Isprs Journal of Photogrammetry and Remote Sensing}, 196:\penalty0 178 -- 196, 2022.

\bibitem[van~der Maaten and Hinton(2008)]{t-SNE}
Laurens van~der Maaten and Geoffrey~E. Hinton.
\newblock Visualizing data using t-sne.
\newblock \emph{Journal of Machine Learning Research}, 9:\penalty0 2579--2605, 2008.

\bibitem[Wang et~al.(2022{\natexlab{a}})Wang, Zhang, Xu, Zhang, Du, Tao, and Zhang]{wang2022advancing}
Di Wang, Qiming Zhang, Yufei Xu, Jing Zhang, Bo Du, Dacheng Tao, and Liangpei Zhang.
\newblock Advancing plain vision transformer toward remote sensing foundation model.
\newblock \emph{IEEE Transactions on Geoscience and Remote Sensing}, 61:\penalty0 1--15, 2022{\natexlab{a}}.

\bibitem[Wang et~al.(2023{\natexlab{a}})Wang, Zhang, Du, Tao, and Zhang]{wang2023scaling}
Di Wang, Jing Zhang, Bo Du, Dacheng Tao, and Liangpei Zhang.
\newblock Scaling-up remote sensing segmentation dataset with segment anything model.
\newblock \emph{Advances in Neural Information Processing Systems}, 2023{\natexlab{a}}.

\bibitem[Wang et~al.(2023{\natexlab{b}})Wang, Albrecht, Braham, Liu, Xiong, and Zhu]{wang2023decur}
Yi Wang, Conrad~M Albrecht, Nassim Ait~Ali Braham, Chenying Liu, Zhitong Xiong, and Xiao~Xiang Zhu.
\newblock Decur: decoupling common \& unique representations for multimodal self-supervision.
\newblock \emph{arXiv preprint arXiv:2309.05300}, 2023{\natexlab{b}}.

\bibitem[Wang et~al.(2023{\natexlab{c}})Wang, Braham, Xiong, Liu, Albrecht, and Zhu]{wang2022ssl4eo}
Yi Wang, Nassim Ait~Ali Braham, Zhitong Xiong, Chenying Liu, Conrad~M Albrecht, and Xiao~Xiang Zhu.
\newblock Ssl4eo-s12: A large-scale multi-modal, multi-temporal dataset for self-supervised learning in earth observation.
\newblock \emph{IEEE Geoscience and Remote Sensing Magazine}, 11\penalty0 (3):\penalty0 98--106, 2023{\natexlab{c}}.

\bibitem[Wang et~al.(2022{\natexlab{b}})Wang, Zeng, Yan, Kang, and Sun]{Polsar}
Zhirui Wang, X.-M. Zeng, Zhiyuan Yan, Jian Kang, and Xian Sun.
\newblock Air-polsar-seg: A large-scale data set for terrain segmentation in complex-scene polsar images.
\newblock \emph{IEEE Journal of Selected Topics in Applied Earth Observations and Remote Sensing}, 15:\penalty0 3830--3841, 2022{\natexlab{b}}.

\bibitem[Wanyan et~al.(2023)Wanyan, Seneviratne, Shen, and Kirley]{wanyan2023dino}
Xinye Wanyan, Sachith Seneviratne, Shuchang Shen, and Michael Kirley.
\newblock Dino-mc: Self-supervised contrastive learning for remote sensing imagery with multi-sized local crops.
\newblock \emph{arXiv preprint arXiv:2303.06670}, 2023.

\bibitem[Waqas~Zamir et~al.(2019)Waqas~Zamir, Arora, Gupta, Khan, Sun, Shahbaz~Khan, Zhu, Shao, Xia, and Bai]{waqas2019isaid}
Syed Waqas~Zamir, Aditya Arora, Akshita Gupta, Salman Khan, Guolei Sun, Fahad Shahbaz~Khan, Fan Zhu, Ling Shao, Gui-Song Xia, and Xiang Bai.
\newblock isaid: A large-scale dataset for instance segmentation in aerial images.
\newblock In \emph{Proceedings of the IEEE/CVF Conference on Computer Vision and Pattern Recognition Workshops}, pages 28--37, 2019.

\bibitem[Wen et~al.(2023)Wen, Cheng, Fang, and Li]{wen2023comprehensive}
Long Wen, Yu Cheng, Yi Fang, and Xinyu Li.
\newblock A comprehensive survey of oriented object detection in remote sensing images.
\newblock \emph{Expert Systems with Applications}, page 119960, 2023.

\bibitem[Wu et~al.(2022)Wu, Liu, Chen, Chen, Dai, and Yuan]{ResidualMoE}
Lemeng Wu, Mengchen Liu, Yinpeng Chen, Dongdong Chen, Xiyang Dai, and Lu Yuan.
\newblock Residual mixture of experts.
\newblock \emph{ArXiv}, abs/2204.09636, 2022.

\bibitem[Xia et~al.(2017)Xia, Hu, Hu, Shi, Bai, Zhong, Zhang, and Lu]{xia2017aid}
Gui-Song Xia, Jingwen Hu, Fan Hu, Baoguang Shi, Xiang Bai, Yanfei Zhong, Liangpei Zhang, and Xiaoqiang Lu.
\newblock Aid: A benchmark data set for performance evaluation of aerial scene classification.
\newblock \emph{IEEE Transactions on Geoscience and Remote Sensing}, 55\penalty0 (7):\penalty0 3965--3981, 2017.

\bibitem[Xiao et~al.(2018)Xiao, Liu, Zhou, Jiang, and Sun]{xiao2018unified}
Tete Xiao, Yingcheng Liu, Bolei Zhou, Yuning Jiang, and Jian Sun.
\newblock Unified perceptual parsing for scene understanding.
\newblock In \emph{Proceedings of the European conference on computer vision (ECCV)}, pages 418--434, 2018.

\bibitem[Xiong et~al.(2024)Xiong, Wang, Zhang, and Zhu]{ofa_net}
Zhitong Xiong, Yi Wang, Fahong Zhang, and Xiao~Xiang Zhu.
\newblock One for all: Toward unified foundation models for earth vision.
\newblock \emph{IGARSS 2024 - 2024 IEEE International Geoscience and Remote Sensing Symposium}, pages 2734--2738, 2024.

\bibitem[Yang et~al.(2021)Yang, Zhang, and Yang]{GASSL}
Longqi Yang, Liangliang Zhang, and Wenjing Yang.
\newblock Graph adversarial self-supervised learning.
\newblock In \emph{Advances in Neural Information Processing Systems}, 2021.

\bibitem[Yuan et~al.(2021)Yuan, Shi, and Gu]{yuan2021review}
Xiaohui Yuan, Jianfang Shi, and Lichuan Gu.
\newblock A review of deep learning methods for semantic segmentation of remote sensing imagery.
\newblock \emph{Expert Systems with Applications}, 169:\penalty0 114417, 2021.

\bibitem[Zhang et~al.(2023)Zhang, Gong, Zhang, Li, Qiao, Ouyang, and Yue]{Meta_trans}
Yiyuan Zhang, Kaixiong Gong, Kaipeng Zhang, Hongsheng Li, Yu Qiao, Wanli Ouyang, and Xiangyu Yue.
\newblock Meta-transformer: A unified framework for multimodal learning.
\newblock \emph{arXiv preprint arXiv:2307.10802}, 2023.

\bibitem[Zhu et~al.(2021)Zhu, Zhu, Li, Wu, Wang, Li, Wang, and Dai]{Uni_Perceiver}
Xizhou Zhu, Jinguo Zhu, Hao Li, Xiaoshi Wu, Xiaogang Wang, Hongsheng Li, Xiaohua Wang, and Jifeng Dai.
\newblock Uni-perceiver: Pre-training unified architecture for generic perception for zero-shot and few-shot tasks.
\newblock \emph{2022 IEEE/CVF Conference on Computer Vision and Pattern Recognition (CVPR)}, pages 16783--16794, 2021.

\bibitem[Zoph et~al.(2022)Zoph, Bello, Kumar, Du, Huang, Dean, Shazeer, and Fedus]{TMoEDS}
Barret Zoph, Irwan Bello, Sameer Kumar, Nan Du, Yanping Huang, Jeff Dean, Noam~M. Shazeer, and William Fedus.
\newblock St-moe: Designing stable and transferable sparse expert models.
\newblock 2022.

\end{thebibliography}
}


\end{document}